\documentclass[journal]{IEEEtran}
\usepackage{balance}
\usepackage{amsmath}
\usepackage{amssymb}
\usepackage{amsthm}

\usepackage{amsfonts}
\usepackage{bm}
\usepackage{subfigure}
\usepackage{booktabs}
\usepackage{tabu}
\usepackage{multirow}
\usepackage{color}
\usepackage[abs]{overpic}
\usepackage{comment}
\usepackage{microtype}
\newcommand{\peicomment}[1]{\textcolor[rgb]{1,0,0} {#1}}
\newcommand{\lincomment}[1]{\textcolor[rgb]{1,0,1} {#1}}
\newcommand{\MR}[1]{\textcolor[rgb]{0,0,0} {#1}}
\ifCLASSINFOpdf
\else
\fi
\begin{document}
%
\title{Pedestrian Detection by Exemplar-Guided Contrastive Learning}
%
%
%


\author{Zebin Lin,
        Wenjie Pei*,
        Fanglin Chen, 
        David Zhang,~\IEEEmembership{Life Fellow,~IEEE} and
        Guangming  Lu*,~\IEEEmembership{Member,~IEEE}
\thanks{*Wenjie Pei and Guangming Lu are corresponding authors.}
\thanks{Zebin Lin, Wenjie Pei, Fanglin Chen and Guangming Lu are with the Department of Computer Science, Harbin Institute of Technology at Shenzhen, Shenzhen 518057, China (e-mail: 19s151113@stu.hit.edu.cn; wenjiecoder@outlook.com; chenfanglin@hit.edu.cn; luguangm@hit.edu.cn).}
\thanks{David Zhang is with the School of Science and Engineering, The Chinese
University of Hong Kong at Shenzhen, Shenzhen 518172, China (e-mail:davidzhang@cuhk.edu.cn)}
\thanks{Manuscript received Xxxx xx, xxxx; revised Xxxx xx, xxxx.}}

\maketitle

\begin{abstract}
Typical methods for pedestrian detection focus on either tackling mutual occlusions between crowded pedestrians, or dealing with the various scales of pedestrians. Detecting pedestrians with substantial appearance diversities such as different pedestrian silhouettes, different viewpoints or different dressing, 
remains a crucial challenge. Instead of learning each of these diverse pedestrian appearance features individually as most existing methods do, we propose to perform contrastive learning to guide the feature learning in such a way that the semantic distance between pedestrians with different appearances in the learned feature space is minimized to eliminate the appearance diversities, whilst the distance between pedestrians and background is maximized. To facilitate the efficiency and effectiveness of contrastive learning, we construct an exemplar dictionary with representative pedestrian appearances as prior knowledge to construct effective contrastive training pairs and thus guide contrastive learning. Besides, the constructed exemplar dictionary is further leveraged to evaluate the quality of pedestrian proposals during inference by measuring the semantic distance between the proposal and the exemplar dictionary. Extensive experiments on both daytime and nighttime pedestrian detection validate the effectiveness of the proposed method.
\end{abstract}


%
\IEEEpeerreviewmaketitle

\section{Introduction}
Pedestrian detection is a challenging task in Computer Vision with many important applications, such as video surveillance~\cite{nascimento2006performance}, driving assistance~\cite{geiger2012we} and  intelligent robotics~\cite{geiger2013vision}.
Deep pedestrian detectors~\cite{ren2015faster,du2017fused,li2017scale,brazil2017illuminating,mao2017can,zhang2017citypersons,wang2018pcn,pang2019mask,song2020progressive,xie2020count,huang2020nms,wu2020temporal}, which benefit from excellent feature learning for images by deep neural networks, have achieved great progress in recent years.

Most existing deep pedestrian detectors view pedestrian detection as a particular case of object detection, and are thus designed following the routine object detection methods. A prominent example is Adapted Faster R-CNN~\cite{zhang2017citypersons}, which directly adapts Faster R-CNN~\cite{ren2016faster} to pedestrian detection. 
Based on such pedestrian detection framework of Adapted Faster R-CNN, many methods are proposed to deal with various challenges for pedestrian detection that are not shared in general object detection. To deal with mutual occlusions between adjacent pedestrians in crowded pedestrian detection scenarios,
Repulsion loss~\cite{wang2018repulsion} is designed to maximize the separation between two adjacent pedestrians. To tackle the same challenge of crowded pedestrian detection, OR-CNN~\cite{zhang2018occlusion} performs body division for pedestrian proposals to highlight the visible body parts while suppressing the occluded parts. Another interesting method for addressing the same problem is Case~\cite{xie2020count}, which proposes a count-and-similarity-aware branch to predict the pedestrian number of a proposal.
Another typical challenge of pedestrian detection is how to deal with various scale (size) of pedestrians, which motivates many research works to address it. Typical methods include ALFNet~\cite{liu2018learning} which is designed based on SSD~\cite{liu2016ssd} and adopts similar multi-stage predictors as Cascade R-CNN~\cite{cai2018cascade}, and SAF R-CNN~\cite{li2017scale} which designs multi-scale detectors to deal with different scales of pedestrians correspondingly.

\begin{figure*}[t]
\centering
    \includegraphics[width=1\linewidth,height=0.3\linewidth]{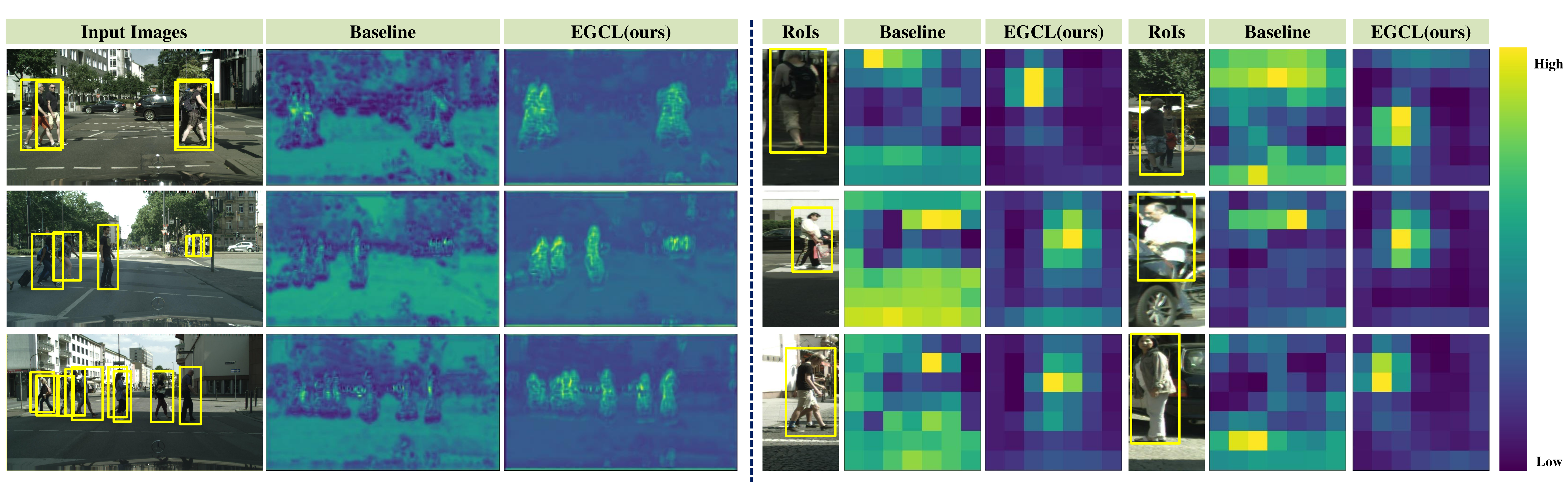}
    \caption{
    Visualization of the feature maps learned by the Baseline model (Adapted Faster R-CNN) and by our \emph{EGCL} model for randomly selected samples from CityPersons~\cite{zhang2017citypersons} validation dataset. \emph{Left}: the feature maps of whole images (C5 block of the features learning head before RPN module) are visualized for both the baseline model and our \emph{EGCL}. \emph{Right}: the feature maps (resized to $7\times 7$) for cropped region proposals (RoIs by RPN module) are visualized for both models. Our \emph{EGCL} is able to detect pedestrians more precisely than the baseline model in both cases since our \emph{EGCL} is designed to maximize the semantic distance between pedestrians and background in the feature space while minimizing the intra-class semantic distance between pedestrians. Note that the Adapted Faster R-CNN is adopted as the baseline model.
    }
\label{Fig:intro}
\end{figure*}


A crucial challenge of pedestrian detection, which has not been addressed well, is to detect pedestrians with a large amount of appearance diversities, especially in large-scale pedestrian detection scenarios. These appearance diversities are potentially resulted from different body silhouettes, different viewpoints, different dressing, different illumination, etc. A robust pedestrian detector should be insensitive to these appearance diversities but focus on the distinction between pedestrians and background. However, most existing methods perform pedestrian detection following the routine way of object detection, and do not explicitly learn to adapt to (ignore) these intra-class appearance differences. As a result, these methods have to allocate much model capacity for learning to recognize each of these diverse appearances (appeared in training data) as individual positive features for pedestrians, which is hardly generalized to large-scale pedestrian detection.

To address above potential limitation, we propose to perform contrastive learning to guide the feature learning in such a way that two objectives are satisfied: 1) the appearance variations between different pedestrians are ignored in the learned feature space, and 2) the semantic distance in the feature space between pedestrians and background is maximized. To this end, we learn a contrastive feature transformation module by contrastive learning, and embed it into typical pedestrian detection frameworks to project the initial feature space into a new feature space in which above two objectives are achieved. Consequently, the pedestrian diversities are eliminated in the feature space and our method can focus on distinguishing between pedestrians and background, namely a binary-classification task, which is the essential of pedestrian detection.

Typically a robust contrastive learning system demands a large amount of training data to be able to generalize to various positive and negative cases. To facilitate the learning efficiency and effectiveness, we construct an exemplar dictionary for pedestrians to be a representative set covering various appearance variations of pedestrians. The obtained exemplar dictionary is not only used to construct effective training set for contrastive learning, but also leveraged to evaluate the quality of the pedestrian proposals during inference by measuring the semantic similarity between proposals and the exemplar dictionary. Such exemplar-contrastive inference is performed jointly with the typical proposal confidence measurement indicated by classification score to achieve more reliable prediction.

Figure~\ref{Fig:intro} presents several real-world examples for pedestrian detection, in which the learned feature maps are visualized for both the baseline model (Adapted Faster R-CNN) and our model. Benefiting from the proposed Exemplar-Guided Contrastive Learning framework (\emph{EGCL}), our model is able to distinguish pedestrians from the background more clearly than the baseline model. To conclude, we make following contributions. 
\begin{itemize}
\item We construct an exemplar dictionary for pedestrians, which comprises representative pedestrian appearances, to facilitate the efficiency and effectiveness of contrastive learning. Furthermore, the constructed exemplar dictionary is also used to refine the confidence score of proposals during inference, based on the our designed metric for semantic distance between proposals and the exemplar dictionary.
\item Based on the exemplar dictionary, we propose an Exemplar-Guided Contrastive Learning framework (\emph{EGCL}) to guide feature learning such that the intra-class semantic distance between pedestrians in the learned feature space is minimized to ignore appearance diversities while the semantic distance between pedestrians and background is maximized. \leavevmode
\item Extensive experiments are conducted on
three typical datasets involving both daytime and nighttime pedestrian detection to validate the effectiveness of our method, both in quantitative and qualitative manners.
\end{itemize}\leavevmode
\vspace{-4pt} 

\section{Related work}

In this section, we firstly make a brief review of pedestrian detection. Then we review the related work on pedestrian detection that focuses on tackling occlusions and dealing with different scales of pedestrians, respectively. Finally, we summarize the methods for contrastive learning briefly.
\subsection{Pedestrian Detection}
\subsubsection{Typical Pedestrian Detection}
Following the routine object detection methods, pedestrian detectors generally consists of three parts, i.e, proposals generation, feature extraction, and classification and regression. Traditional pedestrian detectors utilize the handcrafted features together with downstream classifiers to detect pedestrians. Thus a lot of handcrafted features are proposed to facilitate the separation between pedestrians and background. Dalal et al.~\cite{dalal2005histograms} proposes the Histogram of Oriented Gradients (HOG) feature to reflect the pedestrian's shape and edge information by calculating and integrating the direction and magnitude of the gradient of each pixel. The extracted HOG features are then sent into the downstream classifiers such SVM~\cite{dalal2005histograms} or AdaBoost~\cite{bourdev2005robust} to detect pedestrians. Owing to the excellent representation of objects, the HOG feature makes a great improvement in pedestrian detection and its variants such as Aggregated Channel Features (ACF)~\cite{dollar2014fast},  Integratal Channel Features (ICF)~\cite{dollar2009integral} and Checkerboards~\cite{zhang2015filtered} are proposed successively to detect pedestrians.

As the fast development of deep learning, deep pedestrian detection methods~\cite{wang2018repulsion,li2017scale, wang2018pcn,  zhang2018occlusion,pang2019mask, liu2018learning, xie2020count, huang2020nms} have achieved great progress due to the excellent feature learning capability by deep convolutional networks. Initially, convolutional neural networks (CNN) are directly used to extracted features to replace the traditional handcrafted features~\cite{yang2015convolutional,cao2017learning}. With the great success of Faster R-CNN~\cite{ren2016faster} on object detection, researchers attempt to adapt the Faster R-CNN into pedestrian detection. To address the issue that the downstream classifier of Faster R-CNN degrades the detection performance, RPN+BF~\cite{zhang2016faster} utilizes a boosted forest to substitute for the original classifier of Faster R-CNN and proposes the effective bootstrapping for mining hard negatives. Meanwhile, Adapted Faster R-CNN~\cite{zhang2017citypersons} proposes five improvements to the original Faster R-CNN and exhibits better performance than traditional pedestrian detectors. To eliminate the imbalance between positive samples and negative samples, SDS-RCNN~\cite{brazil2017illuminating} jointly learns pedestrian detection and semantic segmentation by infusing a bounding-box aware semantic segmentation layer into the end of the feature extractor, which enforces the feature focus on the pedestrian's region whilst ignoring the background. Similar to SDS-RCNN, HyperLearner~\cite{mao2017can} integrates the channel feature generated by a semantic segmentation network into the original feature produced by the standard pedestrian detector. Due to the speed advantage of the one-stage object detector, many pedestrian detectors based on one-stage object detection algorithms such as ALFNet~\cite{liu2018learning} and GDFL~\cite{lin2018graininess} are proposed to balance between the detection accuracy and speed of pedestrian detection.

\begin{figure*}[t]
\centering
    \includegraphics[width=1\textwidth,height=0.45\textwidth]{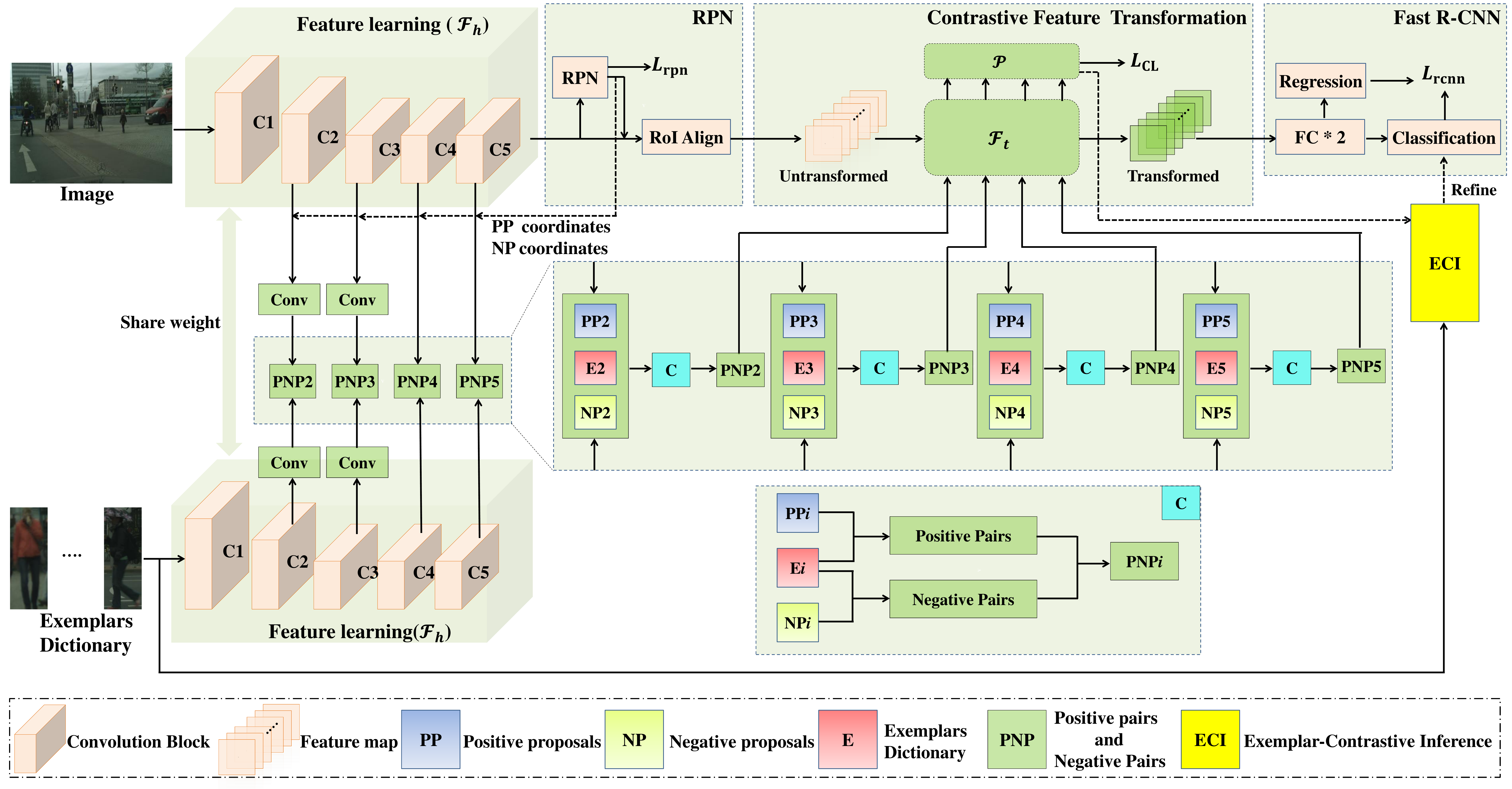}
    \caption{Architecture of the proposed Exemplar-guided contrastive learning network (\emph{EGCL}) for pedestrian detection. Built upon the adapted Faster R-CNN, our \emph{EGCL} performs contrastive learning to learn a feature transformation module $\mathcal{F}_t$ in such a way that the semantic distance between pedestrians in the transformed feature space is minimized whilst the distance between pedestrians and background is maximized. The exemplar dictionary is constructed not only for composing high-quality training pairs for contrastive learning, but also for refining the confidence score of predicted proposals by ECI module.
    }
\label{Fig:Framework}
\end{figure*}

\subsubsection{Occluded Pedestrian Detection}
Occluded pedestrian detection is a challenging task in pedestrian detection due to the information lack of invisible parts of occluded pedestrians and the diversity of occlusions patterns. It aims to solve the problem of mutual occlusions between occluded pedestrians or occlusions caused by background.
The first type of pedestrian detectors for solving occlusions mainly utilize the visible parts of occluded pedestrians or divide occluded pedestrian into multiple parts including visible and invisible parts. Bi-box~\cite{zhou2018bi} pedestrian detector combines the visible part detection with the full-body detection to simultaneously predict the visible part and full-body part of occluded pedestrian. OR-CNN~\cite{zhang2018occlusion} proposes an aggregation loss to encourage positive proposals to be close to their corresponding ground-truth and focuses on highlighting the visible body parts by suppressing the occluded parts. PCN~\cite{wang2018pcn} proposes to predict the score maps of different body parts with a recurrent neural network and designs the context branch for adaptive context selection. Other parts-based methods~\cite{tian2015deep,zhou2017multi} learn a series of part detectors to handle the specific visual patterns of occlusions.

The second type of pedestrian detectors for dealing with occlusions are attention-based methods, which learn robust features with the guidance of the constructed attention maps. Considering that many channels in feature map can be related to different body parts, Zhang et al.~\cite{zhang2018occluded} proposes a channel-wise branch for reweighting the feature map to highlight the visible parts and suppress occluded parts. MGAN~\cite{pang2019mask} introduces a mask-guided attention branch to greatly eliminate the impact of occluded parts whilst highlighting the visible parts. Other pedestrian detectors for solving occlusions include methods based on feature transformation~\cite{zhou2019discriminative}, methods using temporal cue~\cite{wu2020temporal}. 

Both FRCN+A+DT~\cite{zhou2019discriminative} and InterNet~\cite{li2019feature} are proposed to perform feature transformation to improve the performance of pedestrian detection, which is similar to our method. However, our method substantially differs from these two methods.
Our methods differs from FRCN+A+DT~\cite{zhou2019discriminative} in three aspects. Firstly, our method constructs an exemplar dictionary for pedestrians whilst [35] learns one single reference representation for both the pedestrians and the background. Considering the diversity of pedestrian appearance, the exemplar dictionary is able to model such diversities in a more fine-grained manner than the way of learning one single representation. Secondly and most importantly, the whole optimization framework for learning the feature transformation is entirely different between two methods. Our method employs contrastive learning to maximize the margin between positive pairs and negative pairs. In contrast, [35] focuses on minimizing the semantic distance between each sample to the corresponding reference representation. Thirdly, the constructed exemplar dictionary is further leveraged to evaluate the quality of pedestrian proposals during inference by measuring the semantic distance between the proposal and the exemplar dictionary. InterNet~\cite{li2019feature} is designed for object detection. It learns a representative prototype for each class and minimizes the intra-class distance by the proposed interwiner loss. InterNet differs from our method w.r.t. both the target task and the optimization framework.

\subsubsection{Multi-scale Pedestrian Detection}
Detecting different scales of objects is another challenge in object detection. As a particular type of object detection, pedestrian detection also needs to deal with different scales of pedestrians, especially the small-scale pedestrians which tend to be blurred and noisy. Many methods follow divide-and-conquer algorithm to detect different scales of pedestrians. For instance, SAF R-CNN~\cite{li2017scale} trains a large-scale sub-network and a small-scale sub-network to deal with various sizes of pedestrian instances in the image. MS-CNN~\cite{cai2016unified} explores the different depth of feature maps from feature extractor to generate different sizes of proposals, which is followed by a detection network guided by context reasoning. Based on the assumption that different scales of pedestrians bodies can be modeled as 2D Gaussian kernel with various scale variance, TLL~\cite{song2018small} designs a unified fully convolutional network to locate the somatic topological line of pedestrians with line annotation for detecting multi-scale pedestrians. Recently, ascribing the poor performance of detect small-scale pedestrians to the problem of inaccurate location, Cao et al.~\cite{cao2019taking} proposes a location bootstrap module for re-weighting the regression loss. The loss of the predicted bounding box far from the corresponding ground-truth is stressed with high weight while the loss of the predicted bounding box near the corresponding ground-truth is ignored with low weight.

Unlike the aforementioned methods for pedestrian detection, we aim to address the challenge of detecting pedestrians with substantial appearance diversities by performing contrastive learning to guide the feature learning to eliminate the appearance diversities in the learned feature space while maximizing the distance between pedestrians and background.

\subsection{Contrastive Learning}
Contrastive learning aims to guide the feature learning by minimizing the distance between positive pairs and maximizing the distance between negative pairs in the feature space, typically implemented in the form of contrastive loss~\cite{hadsell2006dimensionality,he2020momentum}. One prominent application of contrastive learning is for the self-supervised learning, which seeks to learn robust feature representation from large-scale unlabeled image dataset using a pretext task~\cite{he2020momentum, chen2020simple}.
Self-supervised contrastive learning initially aims to eliminate the performance gap between unsupervised learning and supervised learning in image classification by training different pretext tasks. SimCLR~\cite{chen2020simple} learns effective representation by minimizing the discrepancy between differently augmented views of the same data input. MoCo (v1/v2)~\cite{he2020momentum,chen2020improved} utilizes a momentum-based moving average of the query
encoder to solve the inconsistency of the dictionary keys of negative samples.
Recently, considering that previous works for contrastive learning yield sub-optimal performance when transferred to object detection due to the difference between image classification and object detection, DetCo~\cite{xie2021detco} analyses the essential reasons of inconsistency of classification and detection and introduces instance discrimination as a special pretext task for object detection. Meanwhile, apart from computing contrastive loss in high-level features, DetCo also performs contrastive learning on the low-level features for object detection. 

Contrastive learning framework typically demands a large amount of positive pairs and negative pairs for training~\cite{chen2020simple,chen2020big, he2020momentum,tian2019contrastive,wu2018unsupervised}, which requires a lot of computation resources. To alleviate this problem, we propose a novel exemplar-based offline-online training strategy to train our contrastive learning framework for pedestrians detection, which constructs an exemplar dictionary covering representative pedestrian appearances and thereby utilizes the exemplar dictionary to construct high-quality training pairs for efficient contrastive learning.
\vspace{-4pt}

\section{Exemplar-guided Contrastive Learning}

We aim to optimize feature learning for pedestrian detection in such a way that the distance between pedestrians with various appearances is minimized whilst maximizing the distance between pedestrians and background. To this end, we propose to perform contrastive learning to optimize the feature learning by viewing pedestrian detection as a binary (pedestrian or background) classification problem. To facilitate the efficiency and effectiveness of contrastive learning, we extract 
an exemplar dictionary covering representative pedestrian appearances as prior knowledge to guide the contrastive learning. Besides, the pedestrian exemplar dictionary is 
also leveraged to refine the confidence scores of proposals during inference.


\vspace{-7pt}

\subsection{Pedestrian Detection Framework}
Our proposed contrastive learning framework serves as an auxiliary functional module for optimizing feature learning for pedestrian detection, which can be seamlessly integrated into any existing proposal-based detection framework. As illustrated in Figure~\ref{Fig:Framework}, we build our contrastive learning module upon the Adapted Faster R-CNN~\cite{zhang2017citypersons} as an instantiation.

Adapted Faster R-CNN is a pedestrian detection method designed based on Faster R-CNN~\cite{ren2015faster}. Thus it has similar two-stage modeling process as object detection performed by Faster R-CNN. In the first stage, a Region Proposal Network (RPN) is trained to select a set of high-quality region proposals for potential pedestrians. In the second stage, pedestrian detection is conducted by a classification head and a regression head to predict the class (true or false) of the selected proposals and the corresponding bounding boxes (offsets to the ground-truth), respectively. Two stages are both performed in the deep feature space projected by the feature learning head $\mathcal{F}_h$, which is typically a pre-trained deep neural networks such as VGG-16 ~\cite{simonyan2014very} in Adapted Faster R-CNN.


To detect pedestrians with diverse appearance, we learn a contrastive feature transformation module $\mathcal{F}_t$ by contrastive learning to project the feature space of the feature learning head $\mathcal{F}_h$ into a new feature space. The distance between pedestrians is minimized to eliminate the appearance diversities whilst the distance between pedestrians and background is maximized in this projected feature space.  As shown in Figure~\ref{Fig:Framework}, the contrastive feature transformation module $\mathcal{F}_t$ is embedded between the RPN module and the Fast R-CNN module to perform feature transformation for each of selected region proposals by the RPN. Another reasonable position for $\mathcal{F}_t$ is between the feature learning head $\mathcal{F}_h$ and the RPN, in which case $\mathcal{F}_t$ performs feature transformation on the feature maps for the whole image. We conduct experiments to investigate the effect of the position of $\mathcal{F}_t$ (before or after RPN) on the performance of pedestrian detection in Section~\ref{sec:ablation}.


\vspace{-8pt}
\subsection{Construction of Exemplar Dictionary}
\label{exemplar construction}
The exemplar dictionary is expected to cover full range of appearance diversities of pedestrians. To this end, we perform clustering on the pedestrian images cropped from the training dataset and select the samples closest to each cluster center to compose the exemplar dictionary. Such construction is based on the hypothesis that the pedestrians cropped from a sufficiently large dataset can approximate the pedestrian distribution in real world. 

We employ VGG-16 pre-trained on ImageNet~\cite{deng2009imagenet}, which is the same as feature learning head $\mathcal{F}_h$ in Figure~\ref{Fig:Framework}, to extract features for cropped pedestrian images as input features for clustering. 
Then we perform clustering on the extracted features of all cropped pedestrian images using k-means algorithm~\cite{macqueen1967some} and collect the samples closest to each cluster center to obtain the exemplar dictionary $\mathcal{E}$:
\begin{equation}
    \mathcal{E} = \text{k-means}(\{\mathcal{F}_h(I_i)\};K), \ i= 1, \dots, N,
    \label{eqn:cluster1}
\end{equation}
where $I_i$ is the $i$-th pedestrian image (total $N$ images) cropped from the training set.
Here the number of clusters $K$ is a hyper-parameter to control the size of the constructed exemplar dictionary and balance between training efficiency and effectiveness of contrastive learning. 
Generally, coarse clustering (small $K$) results in a small dictionary with highly representative exemplars, which favors efficient training for contrastive learning but cannot cover full range of pedestrian diversities. In contrast, fine-grained clustering (large $K$) yields a more comprehensive exemplar dictionary but more time is required for contrastive learning to converge. Besides, oversized $K$ may involve low-quality exemplars with low representativeness.

\begin{figure}[!t]
\centering
    \includegraphics[width=1\linewidth]{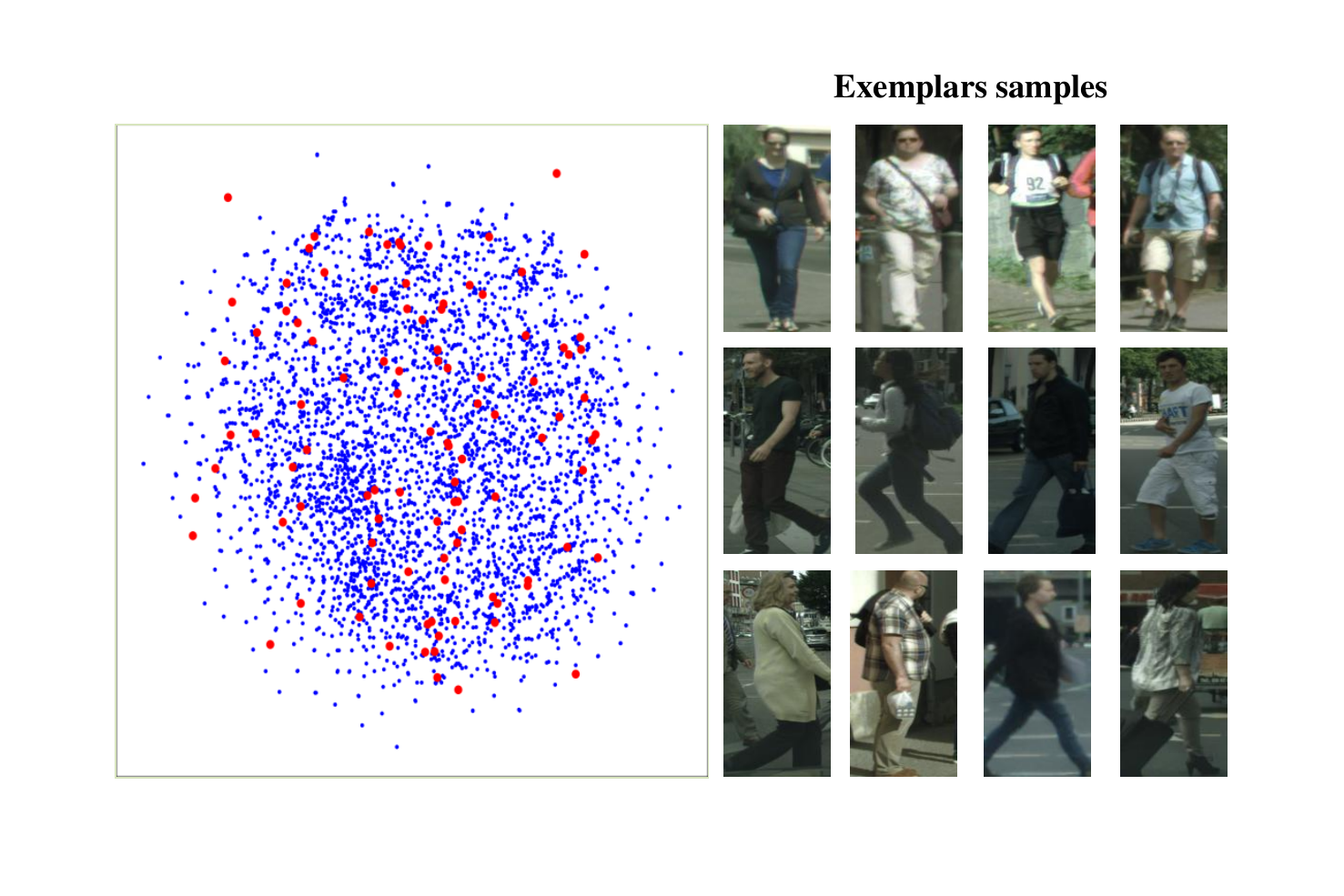}
    \caption{\emph{Left}: t-SNE map of cropped pedestrian images (blue dots) from the training set of CityPersons~\cite{zhang2017citypersons} and the selected samples (red dots) for constructing the exemplar dictionary. \emph{Right}: visualized exemplars randomly selected from the exemplar dictionary.}
\label{Fig:exemplar_tsne}
\end{figure}

Figure~\ref{Fig:exemplar_tsne} presents a t-SNE~\cite{van2008visualizing} map of cropped pedestrian images and the selected samples (red dots) by clustering for constructing the exemplar dictionary, which shows that the selected exemplars span the whole distribution of pedestrian samples to be a representative set of various pedestrians.  Besides, the visualized exemplars, which are randomly selected from the exemplar dictionary, exhibit diverse pedestrian appearances.



\subsection{Multi-level Contrastive Learning}

We conduct multi-level contrastive learning to learn the feature transformation module $\mathcal{F}_t$, utilizing the obtained exemplar dictionary to construct effective training data for contrastive learning.  

\subsubsection{Contrastive learning framework}
\begin{figure}[t]
\centering
    \includegraphics[width=1\linewidth,height=0.5\linewidth]{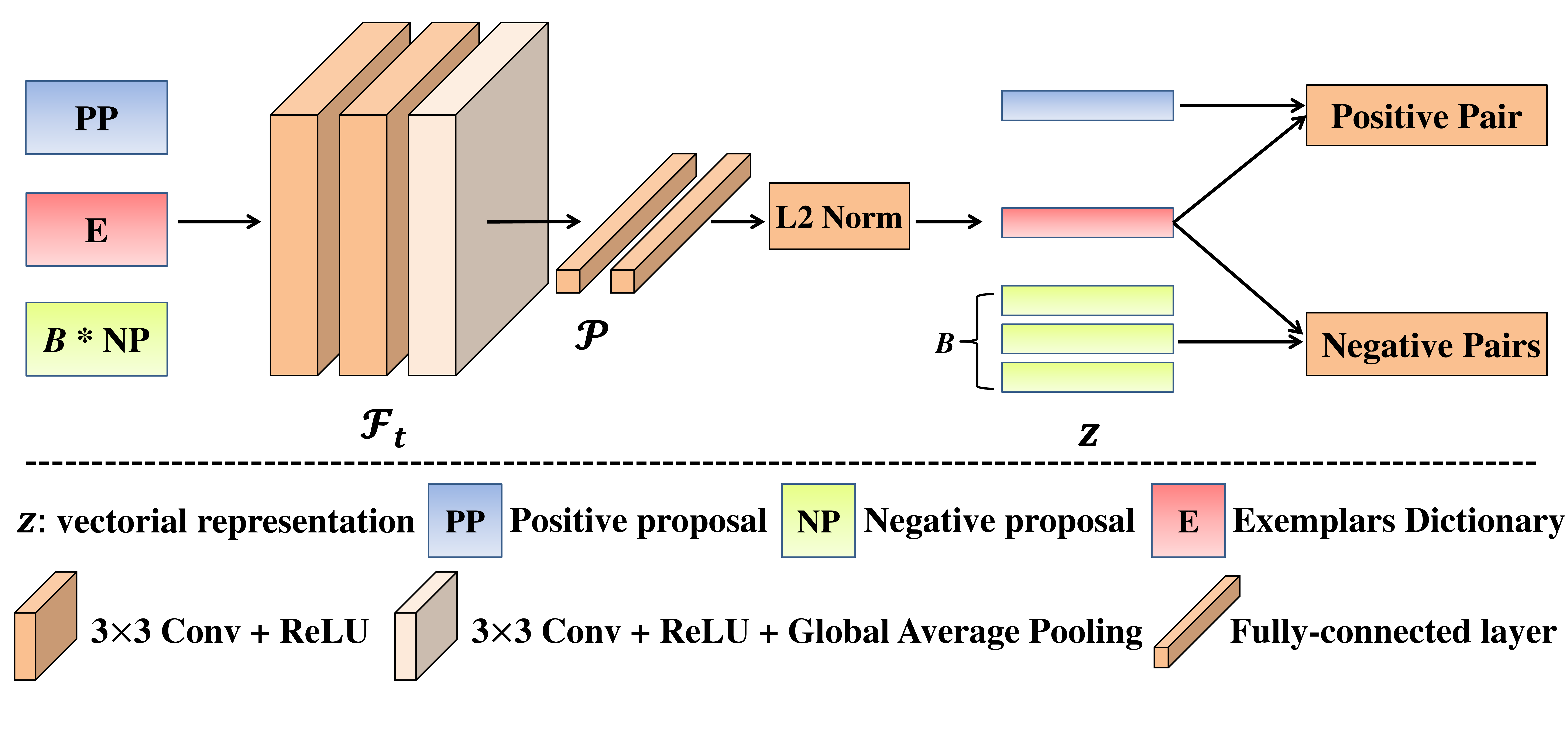}
    \vspace{-20pt}
    \caption{Structure of the contrastive learning framework (CL), where $\mathcal{F}_t$ and $ \mathcal{P} $ refer to the feature transformation module and the projection head, respectively. 
    }
    \vspace{-10pt}
\label{Fig:CL}
\end{figure}

Figure~\ref{Fig:CL} presents the contrastive learning framework, which is a siamese structure consisting of two parts: the feature transformation module $\mathcal{F}_t$ and the projection head $\mathcal{P}$. 
Given a triplet input consisting of three types of image features, namely an exemplar, a positive proposal and $B$ negative proposals, the feature transformation module $\mathcal{F}_t$ transforms all three types of input features into a new feature space while the projection head $\mathcal{P}$ projects the features into a low-dimensional vector by two fully-connected layers. Then the positive proposal and the exemplar compose the positive training pair while the negative proposals and the exemplar compose $B$ negative training pairs for contrastive learning. Note that both positive proposals and negative proposals are obtained by the RPN module based on a predefined IoU overlap threshold. In our implementation, all negative proposals obtained within a batch are used for composing each of the training triplets, thus the value of $B$ is equal to the number of the negative proposals generated in current batch, which could vary in different batches.
Here all input features are extracted from the feature learning head $\mathcal{F}_h$ of the backbone network in Figure~\ref{Fig:Framework}, namely the VGG-16 network. Since the input features are already learned in deep feature space, we design the feature transformation module $\mathcal{F}_t$ as a shallow convolutional network, which is composed of three convolutional layers together with activation layers (ReLU in our implementation). 

We adopt InfoNCE~\cite{oord2018representation}, a widely used contrastive loss function, to supervise our contrastive learning:
\begin{equation}
    \mathcal{L}_{\text{CL}} = -\log \frac{\exp(sim(\mathbf{e},\mathbf{s}_{\text{pos}})/\tau)}{exp(sim(\mathbf{e},\mathbf{s}_{\text{pos}})/\tau) + \sum_{i=1}^{B}\exp(sim(\mathbf{e},\mathbf{s}_{\text{neg}}^{i})/\tau) },
    \label{eqn:cl_loss}
\end{equation}
where $\mathbf{e}$, $\mathbf{s}_{\text{pos}}$, $\mathbf{s}^i_{\text{neg}}$ are the normalized vectorial representations (output from the project head $\mathcal{P}$) of the exemplar, the positive proposal and the $i$-th negative proposal in a triplet input, respectively. 
$sim$ refers to a kernel function measuring the similarity between paired vectors and we opt for dot-product for $sim$ due to its computational efficiency. $\tau$ is a temperature hyper-parameter~\cite{wu2018unsupervised} to tune the sharpness of the exponential function.
Such loss function tries to maximize the similarity of the positive pair while suppressing the similarity values of $B$ negative pairs. Note that the exemplars $\mathbf{e}$ in different triplet input can be different since they are randomly selected from the exemplar dictionary.

\subsubsection{Multi-level contrastive learning mechanism}
A critical issue in object detection is how to deal with different size of objects in a unified framework, which is also challenging in pedestrian detection due to the similar problem formulation.  
To perform contrastive learning that is robust to different size of pedestrians, we perform multi-level contrastive learning by constructing pyramidal feature pairs using different depth of feature maps for a same training pair, either a positive pair or a negative pair. 

Formally, for a training pair consisting a proposal (positive or negative) and an exemplar, we construct multi-level feature pairs by cropping the feature maps for both the proposal and the exemplar from different blocks of the feature learning head $\mathcal{F}_h$. As shown in Figure~\ref{Fig:Framework}, we utilize the features from the $2$-th block to the $5$-th block of $\mathcal{F}_h$ in our implementation. Deeper-level features correspond to larger receptive field in the input image and thus are responsible for detecting larger pedestrians. 
Each level of feature pair has its own contrastive loss, thus the total contrastive loss is defined as:
\begin{equation}
    \mathcal{L}_{\text{CL}} = \sum_{i=2}^5 w_i \mathcal{L}^i_{\text{CL}},
    \label{eqn:total_cl_loss}
\end{equation}
where $w_i, i=2, \dots, 5$ are hyper-parameters for balancing between contrastive losses for different levels of features. In practice, we set $w_2$ to be 1 and tune other hyper-parameters on a held-out small set.


\subsubsection{Offline-online contrastive learning}
To perform contrastive learning effectively, we adopt an offline-online learning strategy: 1) we first train a basic contrastive learner in an offline manner, which can distinguish between pedestrians and background roughly; 2) then we perform online contrastive learning together with the whole pedestrian detection framework, especially focusing on hard-pair contrastive learning for proposals being processed. 

The offline contrastive learning is performed first before the training of the whole pedestrian detection framework. Thus the input features for offline contrastive learning are extracted from the feature learning head $\mathcal{F}_h$ that is only pre-trained on ImageNet without any re-training for pedestrian detection. The training pairs are constructed by randomly selecting cropping pedestrian images as positive proposals and randomly cropped background images as negative proposals from training data. After the offline contrastive learning, the feature transformation module $\mathcal{F}_t$ can roughly perform binary classification between pedestrians and background.

The online contrastive learning is performed after the offline training stage and is conducted jointly with the training of the whole pedestrian detection framework. Thus the parameters of the whole model including the feature learning head $\mathcal{F}_h$, feature transformation module $\mathcal{F}_t$ and other modules are jointly optimized under the supervision of the contrastive learning loss and the losses for pedestrian detection (described subsequently). The positive proposals and negative proposals are all from RPN module on the same image that is being processed for pedestrian detection, which are more challenging to distinguish. Hence, the online contrastive learning further improves the performance of feature transformation module by hard-pair training.
As shown in Figure~\ref{Fig:Framework}, the feature transformation module $\mathcal{F}_t$ is embedded after the RoI Align layer to perform transformation on the features of selected region proposals by the RPN module. Then the transformed proposal features are fed into the Fast R-CNN module to predict the confidence score and the bounding box offsets:
\begin{equation}
    \mathbf{F}_{trans} = \mathcal{F}_t \big[\text{RPN}(\mathcal{F}_h(I))\big],
\end{equation}
where $\mathbf{F}_{trans}$ is the transformed proposal features by $\mathcal{F}_t$ and $I$ is the input image.

\subsection{Collaborative Pedestrian Detection with Exemplar-Contrastive Inference}
The pedestrian detection is typically performed by a classification head and a regression head to predict the confidence score and bounding box given the features of a pedestrian proposal, respectively.  To better evaluate the quality of the proposal, we further leverage the constructed exemplar dictionary to perform exemplar-contrastive inference to measure the semantic distance between the proposal and the exemplar dictionary. 

\begin{figure}[t]
\centering
    \includegraphics[width=1\linewidth]{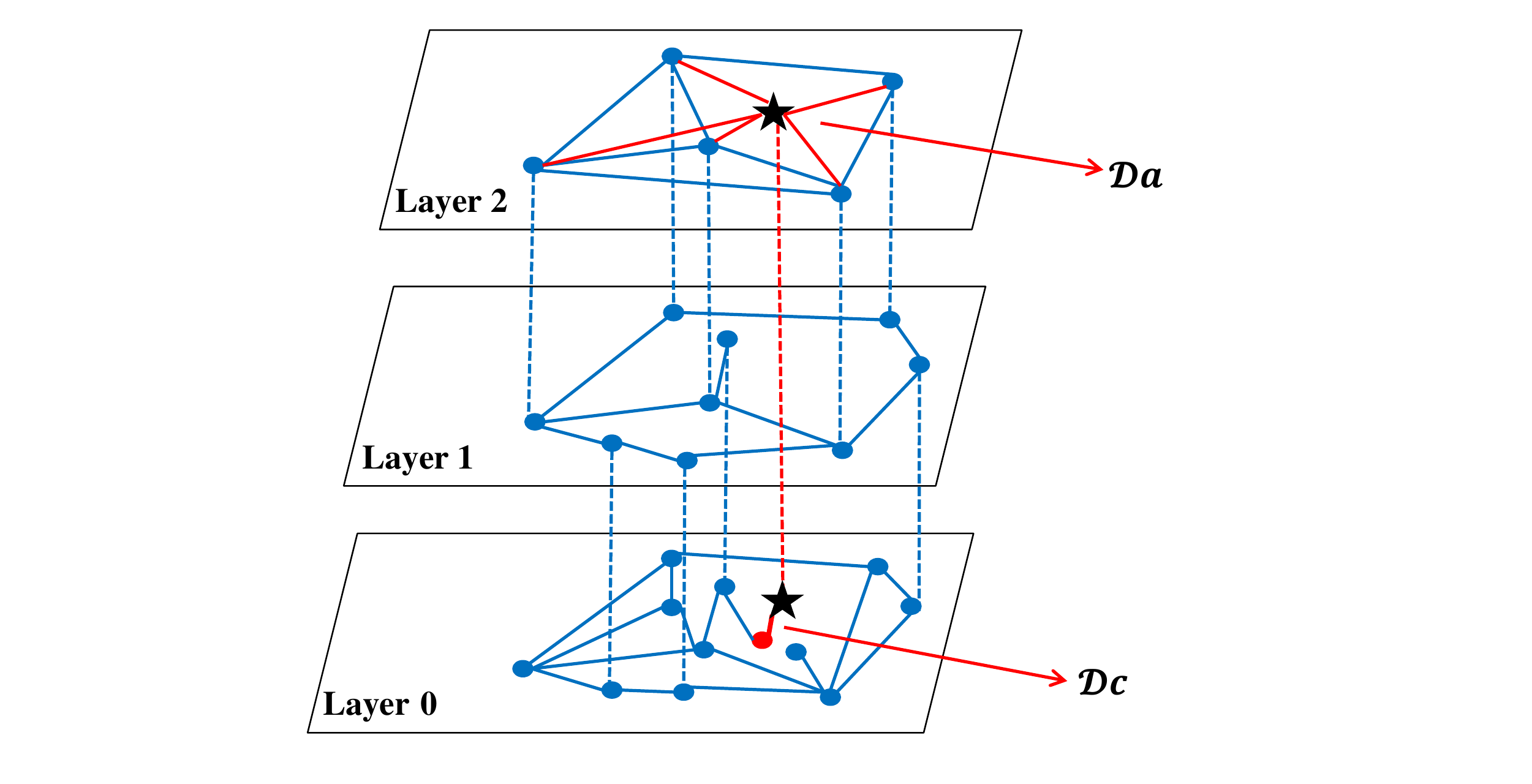}
    \caption{Constructed HNSW graph for the exemplar dictionary. Blue dots refer to exemplars and red dot is the semantically closest exemplar for a proposal denoted by black star. The red solid lines indicate two types of distance $D_a$ and $D_c$.
    }
\label{Fig:hnsw}
\end{figure}

We measure two kinds of semantic distance between a proposal and the exemplar dictionary: 1) the distance between the proposal and the closest exemplar in the dictionary, 2) average distance from the proposal to the whole exemplar dictionary, which is considered to avoid the overfitting of the proposal to a marginal exemplar (in the cases that the closest exemplar happens to be a marginal exemplar). We employ HNSW~\cite{malkov2018efficient} algorithm to build a hierarchical graph in a coarse-to-fine manner for the constructed exemplar dictionary (shown in Figure~\ref{Fig:hnsw}), which is used for efficient navigation to locate the semantically nearest exemplar. The distance $D_c$ between a proposal $\mathbf{s}$ and its nearest exemplar $\mathbf{e}_c$ in the HNSW graph is measured by the opposite similarity value calculated by dot product:
\begin{equation}
    \begin{split}
        & \mathbf{e}_c = \text{HNSW}(\mathcal{E}), \\
        & D_c =  1-\sigma (\mathbf{s} \cdot \mathbf{e}_c),
    \end{split}
\end{equation}
where $\mathcal{E}$ is the constructed exemplar dictionary. Sigmoid function $\sigma$ constrains the value into the interval $[0,1]$. Both $\mathbf{e}_c$ and $\mathbf{s}$ are output features from the projection head $\mathcal{P}$ in the contrastive learning framework. As shown in Figure~\ref{Fig:hnsw}, we measure the average distance $D_a$ between the proposal and the whole exemplar dictionary by:
\begin{equation}
    D_a = \frac{1}{\parallel \text{HNSW}_{\text{top-layer}}(\mathcal{E}) \parallel} \sum_{\mathbf{e} \in \text{HNSW}_{\text{top-layer}}(\mathcal{E})} (1-\sigma (\mathbf{s} \cdot \mathbf{e})).
\end{equation}
Here we approximate the whole distribution of $\mathcal{E}$ by the top layer of HNSW graph with sparse sampling for efficiency. Our method perform exemplar-contrastive inference for evaluating the quality of the proposal  $\mathbf{s}$ by combining two distances $D_c$ and $D_a$. 
Consequently, our method performs collaborative confidence evaluation along with the classification head:
\begin{equation}
    C = (1-\mu -\lambda) P + \mu D_c + \lambda D_a,
    \label{eqn:differ_mu_lamda}
\end{equation}
where $P$ is the predicted confidence by the classification head and $C$ is the final estimated proposal confidence score. $\mu\in [0,1]$ and $\lambda \in [0,1]$ are hyper-parameters to balance between different terms. 

In our implementation, we construct an individual HNSW graph for each level of features for the exemplar dictionary and then select an appropriate HNSW graph according to size of the proposal: the NHSW graph constructed from the deeper feature map are prepared for the larger scale of proposals.

\subsection{End-to-End Parameter Learning}
The contrastive learning framework is first performed in an offline manner under the supervision of contrastive loss in Equation~\ref{eqn:total_cl_loss}, then the online contrastive learning is performed with the whole pedestrian detection framework together, in which the parameters of the whole model are trained in an end-to-end manner. Specifically, the whole model 
is supervised by both the detection loss $\mathcal{L}_{\text{det}}$ and contrastive loss $\mathcal{L}_{\text{CL}}$ jointly:
\begin{align}
\mathcal{L} = \mathcal{L}_{\text{det}} +  \alpha \mathcal{L}_{\text{CL}},
\label{eqn:our-loss}
\end{align}
where $\alpha$ is a balancing weight between two terms. The detection loss $\mathcal{L}_{\text{det}}$ comprises the losses for RPN stage and the losses for Fast R-CNN stage:
\begin{equation}
    \mathcal{L}_{\text{det}} = L_{\text{rpn\_cls}} + L_{\text{rpn\_reg}} + L_{\text{rcnn\_cls}} + L_{\text{rcnn\_reg}}.
\end{equation}
Herein, $L_{\text{rpn\_cls}}$ and $L_{\text{rpn\_reg}}$ refer to the classification loss and regression loss in RPN stage respectively. Similar notations apply to the $L_{\text{rcnn\_cls}}$ and $L_{\text{rcnn\_reg}}$ for Fast R-CNN module.


\section{Experiments}
In this section, we conduct extensive experiments to evaluate the performance of our proposed \emph{EGCL} model on three benchmark datasets including both daytime and nighttime pedestrian detection scenarios. We first perform ablation study to investigate the effectiveness of each key component of our \emph{EGCL}. Then we make qualitative comparison between our \emph{EGCL} and the baseline model (Adapted Faster R-CNN). In the last set of experiments, we compare our model with state-of-the-art methods for pedestrian detection.
\subsection{Experimental Setup}

\subsubsection{Datasets}
We evaluate our proposed method on three standard benchmark datasets including CityPersons~\cite{zhang2017citypersons}, NightOwls~\cite{neumann2018nightowls} and TJU-DHD-pedestrian~\cite{pang2020tju} involving both daytime and nighttime pedestrian detection. CityPersons is collected for daytime pedestrian detection. It comprises 2,975, 500 and 1,525 images for training, validation and test, respectively.  NightOwls is a nighttime pedestrian dataset, in which all the images are captured in the night and dawn time. It contains 128\emph{k} images for training, 51\emph{k} images for validation, and 103\emph{k} images for test. Covering more complex and diverse scenes than CityPersons and NightOwls, TJU-DHD-pedestrian is a challenging dataset for both daytime and nighttime pedestrian detection. It contains 75,246 images with 373,241 labeled pedestrians, which are mixed of daytime and nighttime images. It has two subsets with different scenes, namely  TJU-Ped-campus and TJU-Ped-traffic. Following Pang et al.~\cite{pang2020tju}, we conduct experiments on these two subsets separately.

\subsubsection{Evaluation metrics}
Following the standard pedestrian evaluation protocol~\cite{dollar2011pedestrian}, we choose the log-average Miss Rate over False Positive Per Image (FPPI)  with the range of [
$10^{-2}$,$10^0$] (denoted as $MR^{-2}$) as the evaluation metric. Lower value of $MR^{-2}$ indicates better performance of pedestrian detection. 
For experiments on CityPersons dataset, we report evaluation results on 6 subsets according to the visible ratio (in the area of pedestrian bounding boxes) of each pedestrian: \textbf{R} (reasonable subset with visible ratio in [0.65,1]),  \textbf{HO} (heavy occlusion subset with visible ratio in [0.2,0.65]), \textbf{R+HO} (reasonable and heavy occlusion subset with visible ratio in [0.2, 1]), \textbf{Bare} with visible ratio [0.9, 1.0], \textbf{Partial} with visible ratio [0.65,0.9] and \textbf{Heavy} with visible ratio [0, 0.65].
Following the routine setting~\cite{xie2020count, wu2020temporal}, the pedestrians whose bounding box height are at least 50 pixels are taken for evaluation for CityPersons. On NightOwls dataset, following~\cite{wu2020temporal,neumann2018nightowls} and the official NightOwls evaluation application programming interface (API) \footnote{https://gitlab.com/vgg/nightowlsapi}, we report the evaluation results on the \textbf{Reasonable} subset (non-occluded pedestrians with height $>=$ $50$ pixels), 
\textbf{Reasonable\_small} (non-occluded pedestrians  with height between 50 pixels and 75 pixels),  \textbf{Reasonable\_occ} subsets (occluded pedestrians  with height $>=$ $50$ pixels), and \textbf{All} (pedestrians with height $>=$ 20 pixels). 
On TJU-DHD-pedestrian dataset, we report evaluation results on \textbf{R}, \textbf{HO}, \textbf{R+HO} and \textbf{All} subsets, which is the same with Pang et al.\cite{pang2020tju}.


\subsubsection{Implementation details}
For both datasets, we train our framework on the reasonable subset of training data and evaluate on validation sets (test sets of these datasets are not publicly accessible). The backbone network (VGG-16) of our framework is initialized with ImageNet pre-trained model. We use Adam solver~\cite{kingma2014adam} as optimizer and the mini-batch size is set to 1 image. 
Our framework is trained with 60k steps of gradient descent, in which the initial learning rate of the first 30k steps is set to $1 \times 10^{-4}$ and is further decayed to $1 \times 10^{-5}$ for the left 30k steps. 

\subsection{Ablation Study} 
\label{sec:ablation}
We first perform ablation study to investigate the effectiveness of each key component of our \emph{EGCL} model including the Feature Transformation module (\textbf{FT}), Offline-Online Contrastive Learning (\textbf{OOCL}) and Exemplar-Contrastive Inference (\textbf{ECI}). Specifically, we conduct experiments which begin with the baseline model (Adapted Faster R-CNN) and then incrementally augment the model with each component of \emph{EGCL}. Besides, we also conduct experiments to investigate the effectiveness of the constructed exemplar dictionary. Table~\ref{tab:abaltion_city} and Table~\ref{tab:abaltion_nightowls} present the experimental results of ablation study on CityPersons and NightOwls, respectively. 

\begin{table}[!t]
\centering
\caption{Ablation study of our \emph{EGCL} model on CityPersons validation set in terms of $MR^{-2}$ (lower is better). Baseline refers to Adapted Faster R-CNN. Note that the FT module is embedded before the RPN module in the Asterisked model (Baseline+FT$^*$) while other models mount the FT module after the RPN. For all results, lower is better.}
\renewcommand\arraystretch{2.8} 
\resizebox{1\linewidth}{!}{
\begin{tabular}{l| c  c c c c c c}
\toprule
\multicolumn{1}{l|}{\Large Methods}&\multicolumn{1}{c}{\Large Input Scale} &\multicolumn{1}{c}{\textbf{\Large R}}& \multicolumn{1}{c}{\textbf{\Large HO}}  & \multicolumn{1}{c}{\textbf{\Large R+HO}} & \multicolumn{1}{c}{\textbf{\Large Heavy}} & \multicolumn{1}{c}{\textbf{\Large Partial}} & \multicolumn{1}{c}{\textbf{\Large Bare}}\\
\midrule
\Large Baseline~\cite{zhang2017citypersons}&\Large×1&\Large15.40&\Large64.80&\Large41.45&\Large55.00&\Large18.90&\Large9.30\\
\Large Baseline+FT$^*$&\Large×1&\Large13.79&\Large55.82&\Large31.55&\Large54.44&\Large14.52&\Large8.28\\
\Large Baseline+FT&\Large×1&\Large12.81&\Large53.58&\Large30.57&\Large53.21&\Large14.15&\Large7.28\\
\Large Baseline+FT+OOCL&\Large ×1&\Large12.42&\Large52.25&\Large29.37&\Large52.47&\Large13.21&\Large6.78\\
\Large Baseline+FT+OOCL+ECI (\emph{EGCL})&\Large ×1&\textbf{\Large11.51}&\textbf{\Large50.06}&\textbf{\Large27.93}&\textbf{\Large51.14}&\textbf{\Large11.91}&\textbf{\Large6.15}\\
\bottomrule
\end{tabular}
}
\label{tab:abaltion_city}
\end{table}

\begin{table}[!t]
\centering
\caption{Ablation study of our \emph{EGCL} model on NightOwls validation set in terms of $MR^{-2}$ (lower is better). Baseline refers to Adapted Faster R-CNN. Note that the performance of Baseline model is achieved based on our implementation since no official results are reported. For all results, lower is better.}
\renewcommand\arraystretch{3} 
\resizebox{1\linewidth}{!}{
\begin{tabular}{l| c c c c }
\toprule
\multicolumn{1}{l|}{\LARGE Methods}&\multicolumn{1}{c}{\textbf{\LARGE Reasonable}} &\multicolumn{1}{c}{\textbf{\LARGE Reasonable\_small}} & \multicolumn{1}{c}{\textbf{\LARGE Reasonable\_occ}} & \multicolumn{1}{c}{\textbf{\LARGE All}}\\
\midrule
\LARGE Baseline~\cite{zhang2017citypersons}&\LARGE 18.80& \LARGE26.36&\LARGE58.90&\LARGE 31.45\\
\LARGE Baseline+FT*&\LARGE 17.92&\LARGE 27.08&\LARGE 53.11&\LARGE 30.34 \\
\LARGE Baseline+FT&\LARGE 17.65&\LARGE 26.45&\LARGE 51.46&\LARGE 30.28 \\
\LARGE Baseline+FT+OOCL&\LARGE 16.99&\LARGE 28.05&\LARGE 52.02&\LARGE 29.95\\
\Large Baseline+FT+OOCL+ECI (\emph{EGCL}) &\textbf{\LARGE15.93}&\textbf{\LARGE26.25}&\textbf{\LARGE50.36}&\textbf{\LARGE28.73}\\
\bottomrule
\end{tabular}
}
\label{tab:abaltion_nightowls}
\end{table}

\subsubsection{Comparison with the Baseline (Adapted Faster R-CNN)}
The experimental results in Table~\ref{tab:abaltion_city} and Table~\ref{tab:abaltion_nightowls} show that our \emph{EGCL} model achieves substantial improvement comparing to the Baseline model (Adapted Faster R-CNN) on all subsets of both CityPersons dataset and NightOwls dataset. Considering the performance on \textbf{R} (Reasonable) subset, which is commonly used for performance comparison, \emph{EGCL} boosts the performance of pedestrian detection on  from 15.40 to 11.79 and from 18.80 to 15.93 (in terms of $MR^{-2}$) on CityPersons and NightOwls respectively.


\begin{figure}[t]
\centering
    \includegraphics[width=1.05\linewidth]{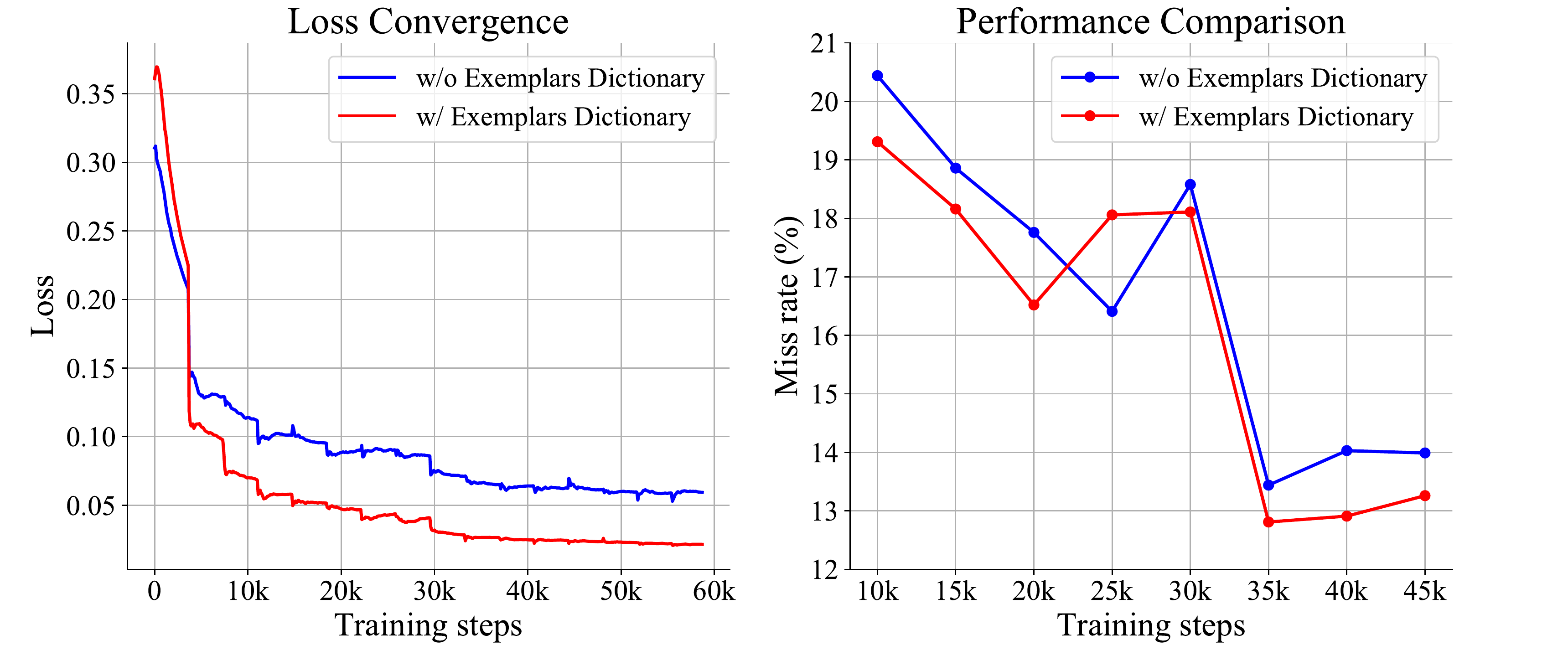}
    \caption{Comparisons between contrastive learning with and without the constructed exemplar dictionary.}
\label{Fig:with_without_exemplars}
\end{figure}
\begin{figure}[t]
\centering
    \includegraphics[width=1\linewidth]{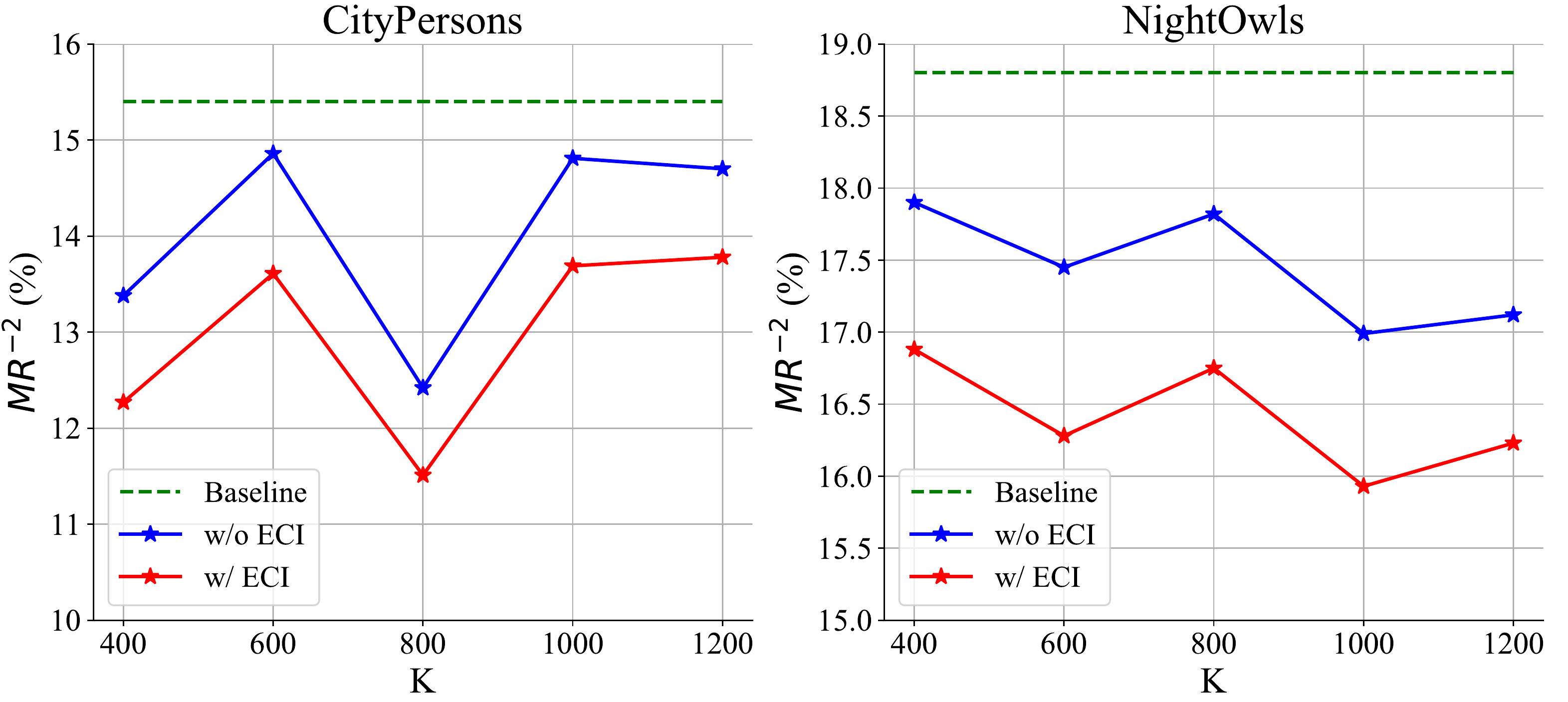}
    \caption{Performance in terms of $MR^{-2}$ of our \emph{EGCL} on \textbf{R} subset, with (red lines) and without (blue lines) the ECI module, as a function of the size of the constructed exemplar dictionary $K$ on the CityPersons and NightOwls dataset, respectively. The performance of the baseline model is also presented for reference.}
\label{Fig:diff_k}
\end{figure}

\begin{figure}[t]
\centering
    \includegraphics[width=1\linewidth]{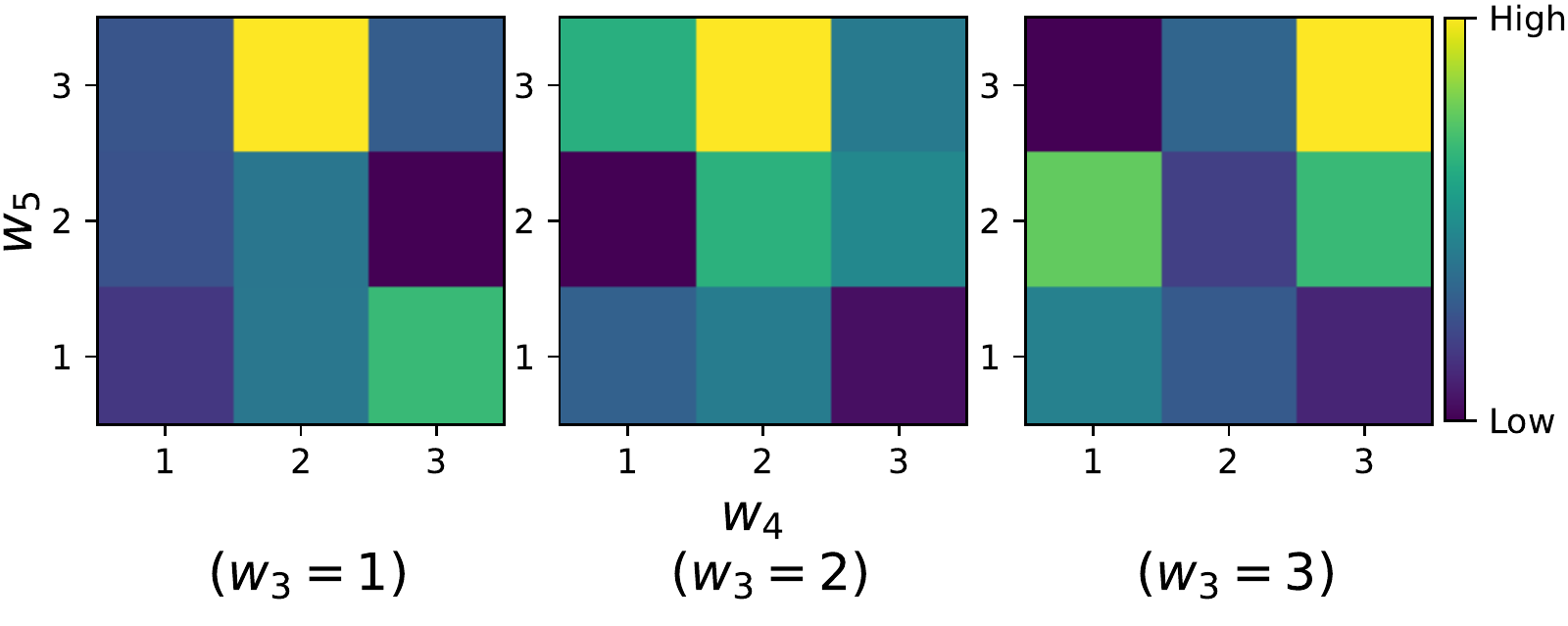}
    \caption{Performance of our EGCL in terms of $MR^{-2}$ on \textbf{R} subset of CityPersons validation dataset as a function of $w_{i}$, i = 3,\dots,5 in Equation \ref{eqn:total_cl_loss} (lower is better).}
\label{Fig:diff_wi}
\end{figure}

\begin{figure}[t]
\centering
    \includegraphics[width=0.8\linewidth]{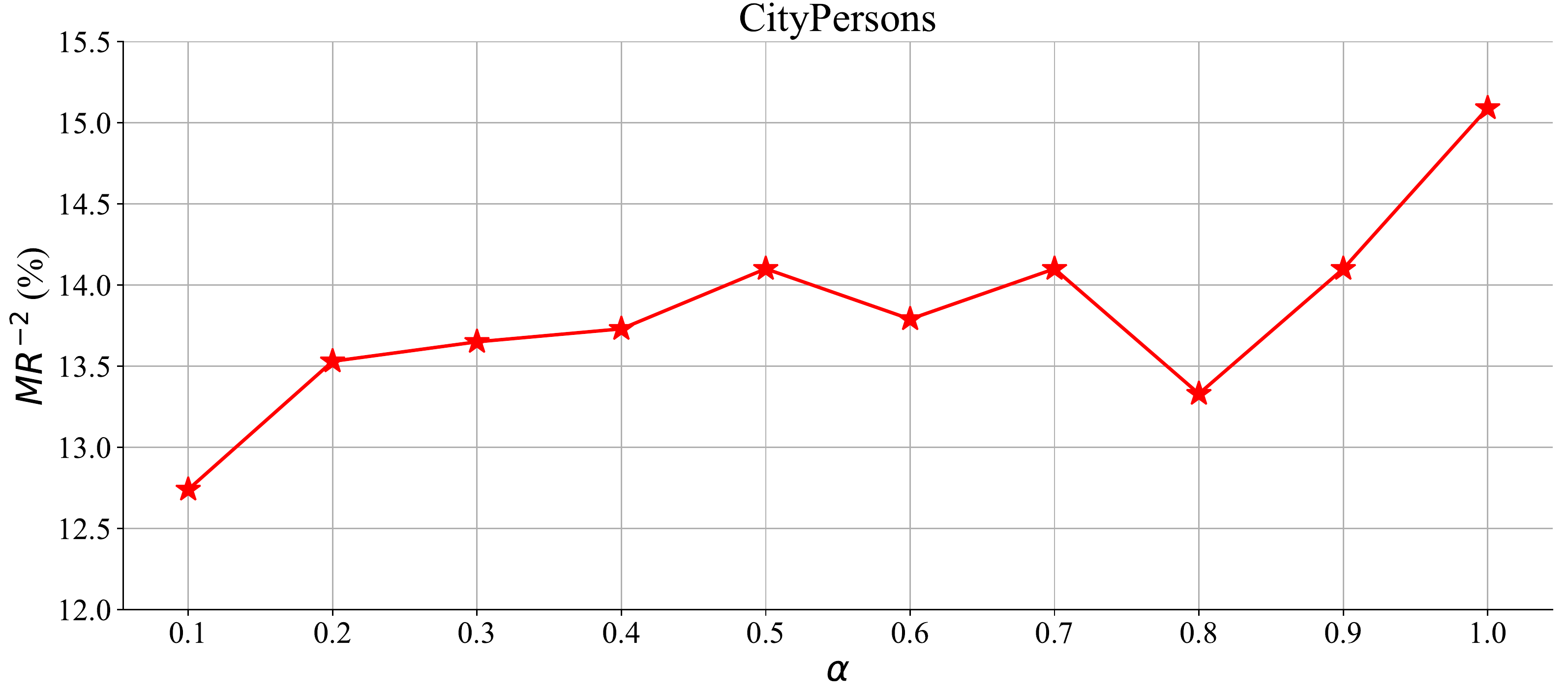}
    \caption{
    Performance of our EGCL in terms of $MR^{-2}$ on CityPersons validation dataset as a function of $\alpha$ in Equation \ref{eqn:our-loss} (lower is better).} 
\label{Fig:diff_alpha}
\end{figure}
\begin{figure}[t]
\centering
    \includegraphics[width=0.6\linewidth]{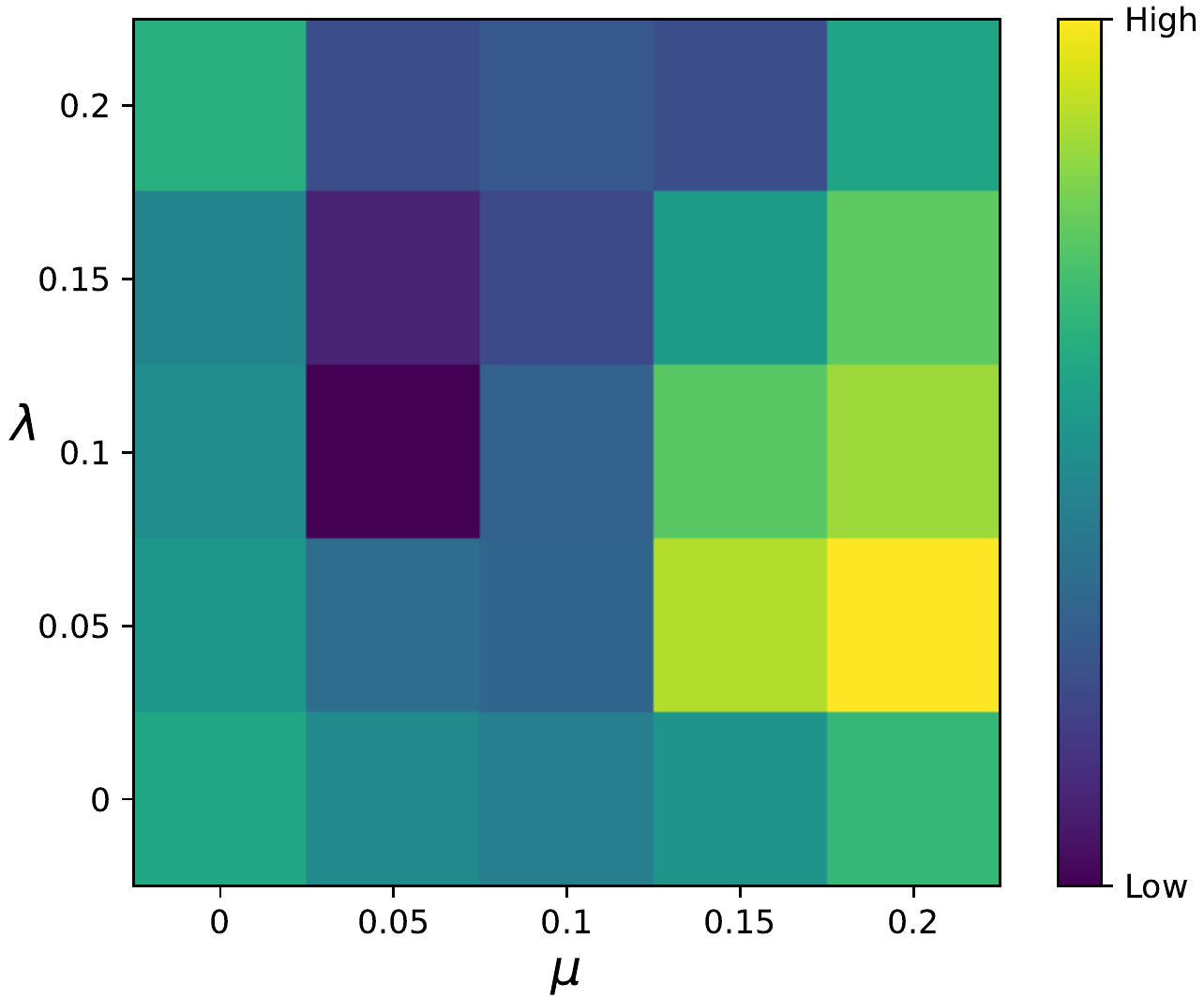}
    \caption{Performance of our EGCL in terms of $MR^{-2}$ on \textbf{R} subset of CityPersons validation dataset as a function of $\mu$  and $\lambda$ in Equation \ref{eqn:differ_mu_lamda} (lower is better).} 
\label{Fig:diff_mu_lambda}
\end{figure}

\subsubsection{Effect of the Feature Transformation module (FT)}
The large performance gap between `Baseline' and `Baseline+FT' on both CityPersons and NightOwls datasets reveals the effectiveness of the feature transformation module of our \emph{EGCL}. Note that we only perform online contrastive learning to train FT module in the setting of `Baseline+FT'. The FT module transforms the initial feature space into a new feature space, in which the semantic distance between pedestrians is minimized while the semantic distance between pedestrians and background is maximized. 

The comparison between `Baseline+FT$^*$' and `Baseline+FT' indicates that embedding the FT module after the RPN module performs better than the other way that before the RPN. We surmise that the FT module placed before the RPN may degenerate the perceptive precision of pedestrians by RPN since the input feature maps of RPN have larger receptive field and thus coarser localization resulted from the convolutional operations in the FT module.

\subsubsection{Effect of Offline-Online Contrastive Learning (OOCL)}
Comparing the performance between `Baseline+FT' and `Baseline+FT+OOCL', we observe that performing offline contrastive learning before online learning can further boost the performance. It validates the theoretical analysis that offline learning is performed to train the feature transformation module (FT) to be a rough classifier between pedestrians and background while online contrastive learning further improves the performance of FT module.

\subsubsection{Effect of Exemplar-Contrastive Inference (ECI)}
The constructed exemplar dictionary is used to not only compose effective training pairs for contrastive learning, but also refine the confidence scores of proposals by proposed exemplar-contrastive inference (ECI). The results in Table~\ref{tab:abaltion_city} and Table~\ref{tab:abaltion_nightowls} validate the effectiveness of ECI on both two datasets.
\begin{figure*}[!t]
\centering
    \includegraphics[width=0.98\linewidth]{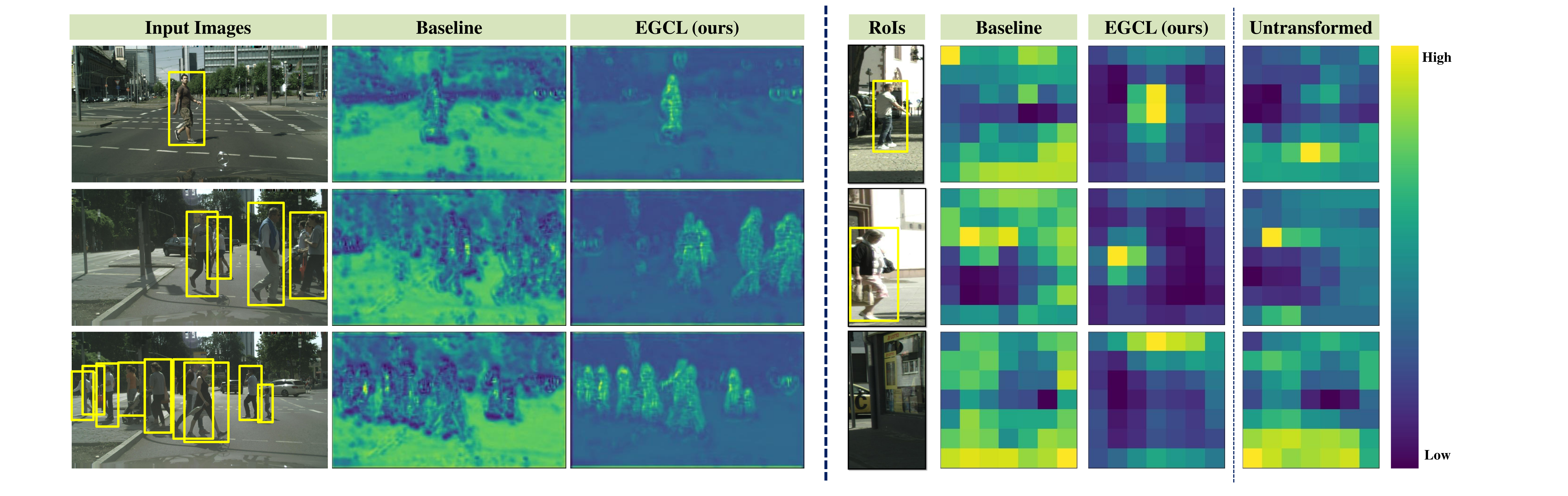}
    \caption{Visualization of the feature maps learned by the Baseline model (Adapted Faster R-CNN) and by our \emph{EGCL} model for randomly selected samples from CityPersons~\cite{zhang2017citypersons} validation dataset. \emph{Left}: the feature maps of whole images (C5 block of the feature learning head $\mathcal{F}_h$ before RPN module) are visualized for both the baseline model and our \emph{EGCL}. \emph{Right}: the feature maps (resized to $7\times 7$) for cropped region proposals (RoIs by RPN module) are visualized for both models. Note that the last sample of RoI visualization is a hard negative sample which tends to be falsely recognized as a pedestrian by the baseline.} 
\label{Fig:qualitative_study_rois_no_ped}
\end{figure*}

\begin{table}[t]
\centering
\caption{Performance of our \emph{EGCL} in terms of $MR^{-2}$ on CityPersons dataset with increasing ratio of occluded samples in the constructed exemplar dictionary (lower is better). }
\renewcommand\arraystretch{1.2} 
\resizebox{0.8\linewidth}{!}{
\begin{tabular}{c c c |c c |c}
\toprule[0.13mm]
\multicolumn{1}{c}{Occluded}&\multicolumn{1}{c}{Total}&\multicolumn{1}{c}{Occluded Ratio} &\multicolumn{1}{|c}{\textbf{HO}} &\multicolumn{1}{c}{\textbf{R+HO}} &\multicolumn{1}{|c}{\textbf{R}}\\
\midrule[0.08mm]
200& 800 &  0.25 & 51.3 & 28.2 & 11.7\\
400& 1000&  0.40 & 50.0 & 27.9 & \textbf{11.5}\\          
800&  1400&  0.57 & \textbf{49.9} & \textbf{27.2} & 12.2\\       
\bottomrule[0.13mm]
\end{tabular}
}
\label{tab:effect_occluded_ratio}
\end{table}

\begin{table}[t]
\centering
\caption{Performance in terms of $MR^{-2}$ on \textbf{R} subset and computational complexity in terms of inference time per image (in second) of our \emph{EGCL} on CityPersons dataset with increasing size of exemplar dictionary $K$. The baseline method and our \emph{EGCL} without ECI module are also presented for reference.}
\renewcommand\arraystretch{1.2} 
\begin{tabular}{l| c c c }
\toprule[0.13mm]
\multicolumn{1}{c}{Methods}&\multicolumn{1}{c}{K}&\multicolumn{1}{c}{Speed}&\multicolumn{1}{c}{\textbf{R}}\\
\midrule[0.08mm]
Baseline& -- & 0.23 &   15.40 \\
\midrule[0.08mm]
w/o ECI &  800 & 0.25 &   12.42  \\  
\midrule[0.08mm]
\multirow{5}*{w/ ECI}& 200  & 0.49  &  12.52  \\   
~& 400  & 0.57 &   12.27\\ 
 ~& 600  & 0.66 &   13.71\\
~& 800  & 0.76 &   11.51\\
~& 1000  & 0.87 &   13.69\\  
\bottomrule[0.13mm]
\end{tabular}
\label{tab:effect_speed_hnsw}
\end{table}



\subsubsection{Effect of the constructed exemplar dictionary}
To investigate the effectiveness of the constructed exemplar dictionary, we compare the performance of contrastive learning using and not using the exemplar dictionary for constructing the training pairs. Figure~\ref{Fig:with_without_exemplars} illustrates the comparison between two ways w.r.t. the loss convergence and performance respectively. Using the constructed exemplar dictionary, the loss of contrastive learning converges faster and reaches a lower value than that without exemplar dictionary. Further, exemplar dictionary also empowers contrastive learning to achieve a better performance in terms of $MR^{-2}$, which demonstrates the effectiveness of exemplar dictionary for contrastive learning.

\noindent\textbf{Effect of the size of exemplar dictionary $K$.} The size of the exemplar dictionary $K$ is a hyper-parameter to be tuned in our experiments. Typically, small $K$ leads to an exemplar dictionary which has highly representative exemplars but cannot cover full range of pedestrian diversities. In contrast, oversized $K$ may involve low-quality exemplars with low representativeness. Figure~\ref{Fig:diff_k} presents the performance of our \emph{EGCL} as a function of $K$ on CityPersons and NightOwls datasets respectively. The experiments are conducted under two different settings: 1) the ECI module is unmounted, under which setting the effect of different $K$ on sole contrastive learning is evaluated; 2) the ECI module is activated and the experiments are designed to investigate the effect of different $K$ on both contrastive learning and the exemplar-contrastive inference. We make following observations from the results. 1) Under both settings, the performance of our model is being improved at the beginning as the increase of $K$ and reaches an optimal point, then the performance begins to degrade when provided larger size of exemplar dictionary. These results are consistent with our theoretical analysis. 2) The ECI module consistently improves the performance of our model when setting $K$ to different values, which indicates the robustness of the proposed ECI module. 3) Though the performance of our model fluctuates with varying values of $K$, the performance is always better than the baseline model (\emph{Adapted Faster R-CNN)}, which again reveals the effectiveness of the ECI module.

Since the `reasonable' samples account for the vast majority of pedestrians in the training data and real-world scenarios as well, the exemplar dictionary contains mostly 'reasonable' exemplars and only 20\% occluded exemplars if it is constructed uniformly based on the training data. However, dealing with the heavy-occluded pedestrians remains a crucial challenge in pedestrian detection. To improve the performance of our \emph{EGCL} in detecting the heavy-occluded pedestrians, we increases the ratio of occluded samples in the exemplar dictionary by simply replicating the existing occluded exemplars. Table~\ref{tab:effect_occluded_ratio} shows the performance of our \emph{EGCL} on CityPersons dataset with increasing ratio of occluded samples in the exemplar dictionary. We observe that increasing the ratio of the occluded exemplars indeed improves the performance of our \emph{EGCL} to detect the heavy-occluded pedestrians (`\textbf{HO}' and `\textbf{R+HO}' subsets). However, too large ratio of the occluded exemplars results in the performance degradation of our model on the `Reasonable' subset (\textbf{R}). 


\subsubsection{\textbf{Tuning of hyper-parameters}}
To illustrate the effect of the hyper-parameters involved in our method on the performance, we report the performance of our method with different settings of hyper-parameters. To be specific, Figure~\ref{Fig:diff_wi} visualizes the distribution map of performance as a function of $w_i, i=3, \dots, 5$ in Equation~\ref{eqn:total_cl_loss} when performing grid search for 3 parameters. Note that $w_2$ is fixed to be 1 to tune other balancing weights. Figure~\ref{Fig:diff_alpha} shows the effect of $\alpha$ in Equation~\ref{eqn:our-loss} on the performance while Figure~\ref{Fig:diff_mu_lambda} presents the performance of our model when varying the values of $\mu$ and $\lambda$ in Equation~\ref{eqn:differ_mu_lamda} during grid search. In practice, we tune all these hyper-parameters on a held-out small set split from the training set, on which we can obtain the consistent correlations between the performance of our model and the settings of these hyper-parameters.

\subsubsection{\textbf{Computational complexity}} 
To analyze the computational complexity of our model, we measure the inference time of our model with different configurations in Table~\ref{tab:effect_speed_hnsw}. Comparing the inference time between the baseline and our model without ECI, we observe that the contrastive learning incurs little extra time while yielding substantial performance improvement. On the other hand, ECI module is distinctly more time-consuming than the contrastive learning. The time cost of the ECI module increases roughly linearly with the size of exemplar dictionary $K$.

\begin{figure*}[!t]
\centering
    \includegraphics[width=0.9\linewidth]{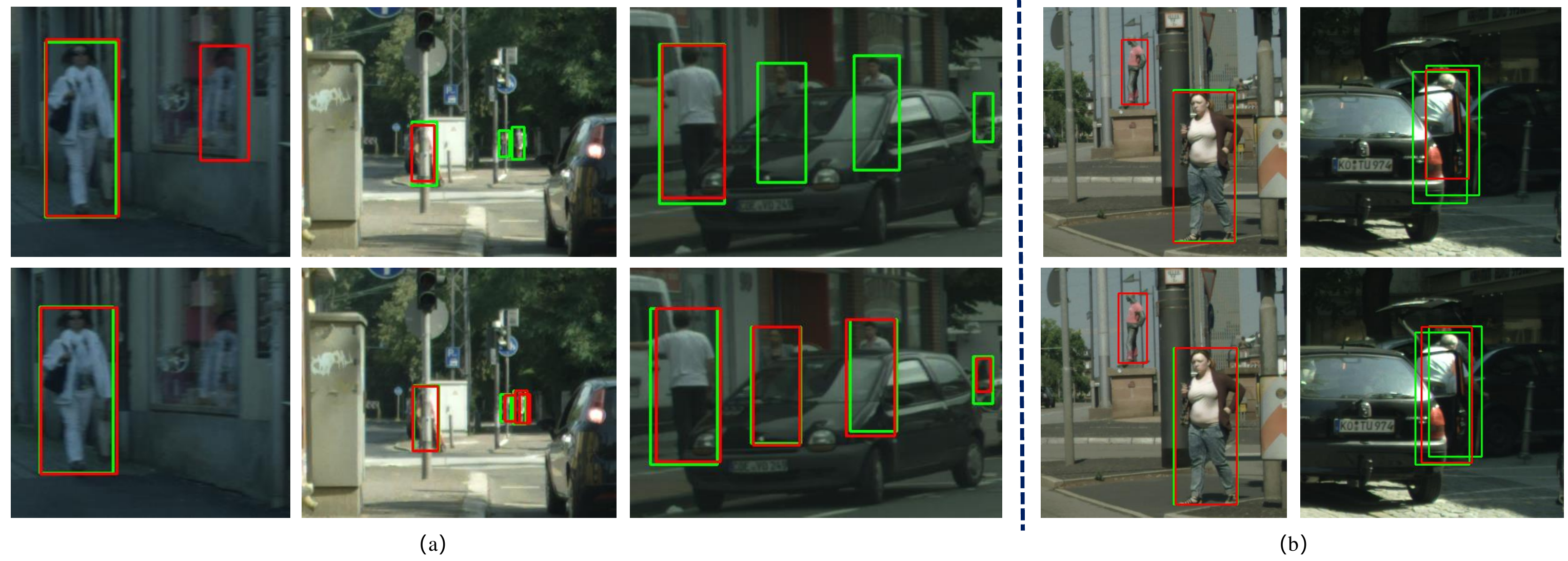}
    \caption{Visualization of detection results by our model and the baseline. (a) Our model is able to detect all pedestrians correctly whilst the baseline cannot recognize the pedestrians with small-scale or heavily occluded pedestrians. (b) Two challenging examples on which both our model and the baseline fail to detect pedestrians correctly.}
\label{Fig:visualization_results}
\end{figure*}
\begin{figure}[!t]
\centering
    \includegraphics[width=1\linewidth]{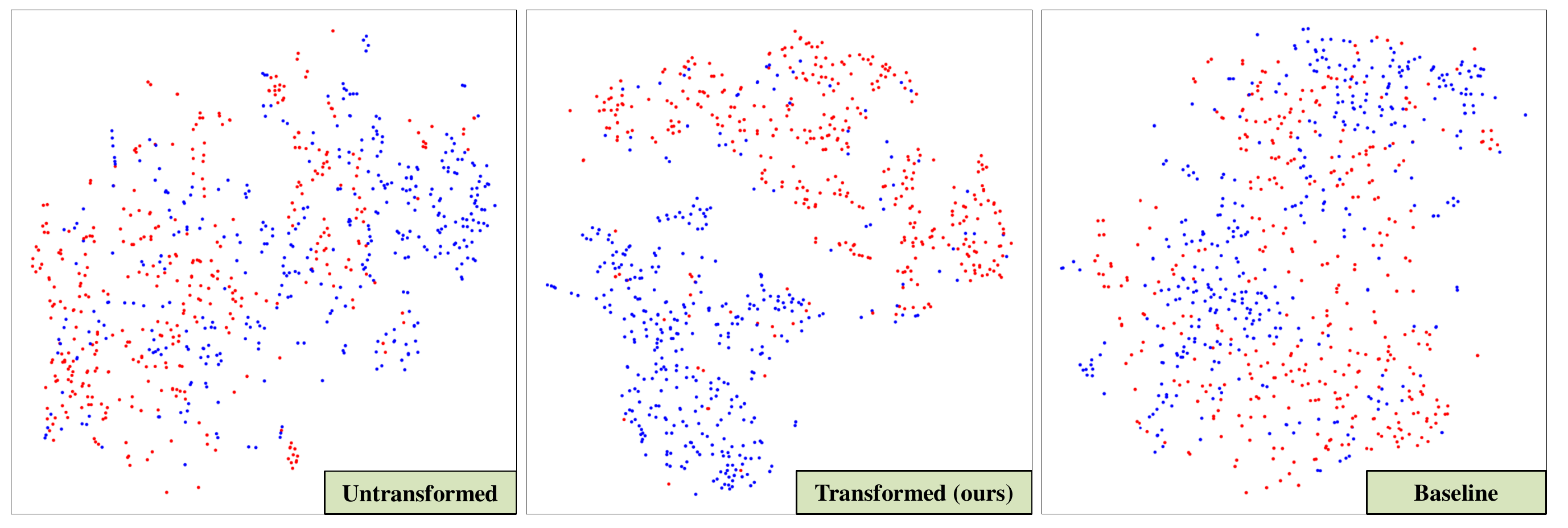}
    \caption{The t-SNE map of RoI features by different models, in which the blue dots and the red dots refer to the positive proposals (IoU $>$ 0.6) and negative proposals (IoU $<$ 0.4). \emph{Left} and \emph{Middle:} RoI features before and after the feature transformation module (FT) of our trained \emph{EGCL} respectively; \emph{Right:} RoI features of the baseline model.} 
\label{Fig:ped_back_tsne}
\end{figure}

\subsection{Qualitative Study on Contrastive learning}
\subsubsection{Visualization of learned feature maps} To obtain more insights into the effectiveness of our \emph{EGCL}, we make a qualitative comparison between our \emph{EGCL} and the baseline model (Adapted Faster R-CNN)  by visualizing the feature maps of both models. Figure~\ref{Fig:qualitative_study_rois_no_ped} visualizes both the feature maps for the whole images (C5 block of feature learning head $\mathcal{F}_h$) and that for cropped region proposals (RoIs). Our \emph{EGCL} is able to detect pedestrians more precisely since it minimizes the semantic difference between pedestrians and maximizes the distance between pedestrians and background in the feature space. Besides, the untransformed feature maps for RoIs before the feature transformation module of our \emph{EGCL} are also visualized in Figure~\ref{Fig:qualitative_study_rois_no_ped} for reference. An interesting observation is that  our model is able to correctly deny the confusing (hard) negative sample in the last row of RoI visualization, which tends to be falsely recognized as a pedestrian by the baseline and the untransformed feature maps of our model.

\subsubsection{Separability of RoI features} We further apply t-SNE~\cite{van2008visualizing} to RoI features by different models to compare the their capability to distinguish between positive proposals (IoU $>$ 0.6) and hard negative proposals (IoU $<$ 0.4). Figure~\ref{Fig:ped_back_tsne} presents t-SNE maps for the baseline model and our model (including two cases: before and after the feature transformation module (FT)). The t-SNE map of RoI features after the FT module by our \emph{EGCL} shows more separability than other two t-SNE maps.

\subsubsection{Visualization of pedestrian detection}
We visualize the results of pedestrian detection by our \emph{EGCL} and the baseline model in Figure~\ref{Fig:visualization_results} to have a qualitative comparison. We first present the detection results of both models on three examples that are randomly selected from the training set in Figure~\ref{Fig:visualization_results} (a). The results show that our model is able to detect all pedestrians precisely, even those with quite small size (in the second sample) or occluded heavily (in the third sample) which are missed by the baseline model. Besides, the baseline falsely captures the reflection of the pedestrian in the mirror (in the first sample) while our model can discriminate the reflection from real pedestrians.

\noindent\textbf{Limitations.} Figure~\ref{Fig:visualization_results} (b) presents two challenging examples, on which both our model and the baseline fails to detect pedestrians correctly. In the first sample, our \emph{EGCL} mis-recognizes a vivid human statue as a pedestrian. In this case, it is quite challenging to distinguish it from genuine pedestrians relying on sole statue features. A potential solution that can be explored in the future work is to model the semantic correlation between the statue and its surrounding objects like stone pedestal, and thereby make inference more correctly. In the second example, there are two pedestrians which are highly overlapped. Our model cannot detect the occluded pedestrian probably due to the removal of highly overlapped proposals by NMS algorithm. Those two examples reveals two potential limitations of our model, which we intend to address in the future work.
\begin{table*}[!t]

\centering
\caption{Performance (in terms of $MR^{-2}$) of our \emph{EGCL} and other methods on CityPersons validation subset (lower is better). To have a fair comparison, the experiments are conducted in three different settings w.r.t. the used backbone and the scale of input images.}
\renewcommand\arraystretch{1.5} 
\begin{tabular}{l| c c c c c c c c}
\toprule
\multicolumn{1}{l|}{Methods} &\multicolumn{1}{c}{Backbone} &\multicolumn{1}{c}{Input Scale} &\multicolumn{1}{c}{\textbf{R}}& \multicolumn{1}{c}{\textbf{HO}}  & \multicolumn{1}{c}{\textbf{R+HO}} & \multicolumn{1}{c}{\textbf{Heavy}} & \multicolumn{1}{c}{\textbf{Partial}} & \multicolumn{1}{c}{\textbf{Bare}}\\
\midrule
F.RCNN+ATT-vbb~\cite{zhang2018occluded}  & VGG-16 & ×1 &  16.4   &  57.3     &   --  & -- & -- & -- \\
F.RCNN+ATT-part~\cite{zhang2018occluded}  & VGG-16 & ×1 &  16.0   &  56.7     &   38.2  & -- & -- & -- \\
Adapted FasterRCNN~\cite{zhang2017citypersons}  & VGG-16        & ×1          &  15.4                                &     64.8                             &   41.5      & 55.0 &18.9 &  9.3                            \\ 
OR-CNN~\cite{zhang2018occlusion} & VGG-16           & ×1        &  12.8   &    55.7     &   --   & 55.7 & 15.3 &  6.7                            \\ 
Adaptive-NMS~\cite{liu2019adaptive} & VGG-16   & ×1        &   11.9         &        55.2        &  --      & 55.2 &12.6 & 6.2                              \\ 
MGAN~\cite{pang2019mask} & VGG-16   & ×1        &   11.5         &        51.7        &  --      & -- &-- & --                             \\ 
HGPD~\cite{li2020learning} & VGG-16   & ×1        &   11.3         &        51.7        &  --      & -- &-- & --                             \\
$R^{2}$NMS~\cite{huang2020nms} & VGG-16   & ×1        &   11.1         &        53.3      &  --      & -- &-- & --                             \\
Case~\cite{xie2020count} & VGG-16   & ×1        &   \textbf{11.0}         &        50.3      &  --      &  --      & --      &  --                                 \\
\emph{EGCL} (ours) & VGG-16  & ×1        & 11.5                                 &  \textbf{50.0}                                &   \textbf{27.9}      & \textbf{51.1}& \textbf{11.9} &  \textbf{6.1}  \\
\midrule
InterNet~\cite{li2019feature}& ResNet-50&×1& 17.9&70.2 & 48.2& 60.4 & 23.7& 13.5
 \\
TLL~\cite{song2018small} & ResNet-50        & ×1          &  15.5                                &    --                          &   --      & 53.6 &17.2 &  10.0                            \\ 
TLL+MRF~\cite{song2018small} &  ResNet-50    & ×1          &  14.4                               &      52.0                         &  -- & 52.0 & 15.9 & 9.2                               \\
Repulsion Loss~\cite{wang2018repulsion}  &  ResNet-50  & ×1          &     13.2                            &      56.9                           &   --   & 56.9 & 16.8 & 7.6                                 \\ 
ALFNet~\cite{liu2018learning} &  ResNet-50        & ×1      &    12.0        &    43.8      &  26.3     & 51.9 & \textbf{11.4} &    8.4                             \\ 
CrowdDet~\cite{chu2020detection} & ResNet-50  & ×1        & 12.1                                &  40.0                                &  25.4     & 47.7 & 12.9 &  7.5 \\
CrowdDet+\emph{EGCL} (ours) & ResNet-50  & ×1        & \textbf{10.9}                                 &  \textbf{39.3}                              &   \textbf{24.8}    & \textbf{46.4} & 11.6 &  \textbf{7.4}  \\
\midrule
FRCN+A+DT~\cite{zhou2019discriminative}&VGG-16& ×1.3& 11.1& --& --&\textbf{44.3} & 11.2 & 6.9 \\
NOH-NMS~\cite{zhou2020noh} & ResNet-50 & ×1.3 &10.8&--&--&53.0&11.2&\textbf{6.6} \\
CrowdDet~\cite{chu2020detection} & ResNet-50 & ×1.3 &10.7&38.0&24.3&45.8&10.7&7.3 \\
MGAN~\cite{pang2019mask}& VGG-16& ×1.3& 10.3 & 49.6 & -- & -- & -- & --\\
Case~\cite{xie2020count} & VGG-16   & ×1.3        &   \textbf{9.6}        &        48.2      &  --      & -- &-- & -- \\
CrowdDet+\emph{EGCL} (ours) & ResNet-50 & ×1.3 & 10.5&\textbf{37.2}&\textbf{23.8}&45.3&\textbf{10.2}&6.8\\
\bottomrule
\end{tabular}
\label{tab:comparison_city}
\end{table*}

\begin{table}[!t]
\centering
\caption{Performance ( in terms of $MR^{-2}$) of our \emph{EGCL} and other methods on NightOwls validation subset (lower is better).}
\renewcommand\arraystretch{1.2} 
\resizebox{0.8\linewidth}{!}{
\begin{tabular}{l| c c }
\toprule[0.13mm]
\multicolumn{1}{l|}{Methods}&\multicolumn{1}{c}{Backbone} &\multicolumn{1}{c}{\textbf{Reasonable}} \\
\midrule[0.08mm]
ACF~\cite{dollar2014fast} &  -- &  51.68\\
Checkerboards~\cite{zhang2015filtered}  &  -- & 39.67 \\
Vanilla Faster R-CNN~\cite{ren2016faster} & VGG-16 &  20.00 \\
 Adapted Faster R-CNN~\cite{zhang2017citypersons} & VGG-16& 18.81 \\
 RPN+BF~\cite{zhang2016faster}   &  VGG-16          &          23.26    \\
 SDS-RCNN~\cite{brazil2017illuminating}    &  VGG-16          &     17.80  \\        
 \MR{TFAN~\cite{wu2020temporal}}    & \MR{ResNet-50}         &     \MR{16.50} \\
 \emph{EGCL} (ours)   &  VGG-16         & \textbf{ 15.93}\\                               
\bottomrule[0.13mm]
\end{tabular}
}
\label{tab:comparison_nightowls}
\end{table}




\vspace{-5pt}
\subsection{Comparison with State-of-the-art methods}

\subsubsection{CityPersons dataset}
To evaluate the performance of our \emph{EGCL} on daytime pedestrian detection, we compare our \emph{EGCL} model with other state-of-the-art methods for pedestrian detection on CityPersons dataset. These methods include F.RCNN+ATT-vbb~\cite{zhang2018occluded}, F.RCNN+ATT-part~\cite{zhang2018occluded}, Adapted Faster R-CNN~\cite{zhang2017citypersons}, OR-CNN~\cite{zhang2018occlusion}, Adaptive-NMS~\cite{liu2019adaptive}, MGAN~\cite{pang2019mask}, HGPD~\cite{li2020learning}, TLL~\cite{song2018small}, $R^{2}$NMS~\cite{huang2020nms}, TLL+MRF~\cite{song2018small}, Replusion loss~\cite{wang2018repulsion}, ALFNet~\cite{liu2018learning}, NOH-NMS~\cite{zhou2020noh}, \MR{FRCN+A+DT~\cite{zhou2019discriminative}},Case~\cite{xie2020count} and CrowdDet~\cite{chu2020detection}. We also adapts InterNet~\cite{li2019feature} from object detection to pedestrian detection and evaluate its performance.
Note that these state-of-the-art pedestrian detectors are trained on different backbones (VGG16 or ResNet-50) for feature learning head and different input scales ($\times 1$ or $\times 1.3$). For a fair comparison, as shown in Table~\ref{tab:comparison_city}, we compare our methods with these state-of-the-art methods in three settings w.r.t. the adopted backbone and the input scale. Using $\times 1$ scale of input images with VGG16 as the backbone, our \emph{EGCL} achieves the best performance on most of subsets and performs on par with the first-rank method on \textbf{R} subset. In particular, our method outperforms the second place substantially on both \textbf{HO} and \textbf{R+HO} subsets, which reveals the advantages of our model in detecting heavy-occluded pedestrians. In the setting of adopting ResNet-50 as the backbone and using $\times 1$ scale of input images, our \emph{EGCL} performs best on all subsets except \textbf{Partial} subset. It is worth noting that our \emph{EGCL} outperforms the second place by $1.2\%$ on  \textbf{R} subset, which is the most important evaluating metric for pedestrian detection. Our \emph{EGCL} also performs well in the setting of using $\times 1.3$ scale of input image and ResNet-50 as backbone. These favorable results validate the effectiveness of our model.


\subsubsection{NightOwls dataset}
Detecting pedestrians in nighttime is more challenging than in daytime due to difficult discrimination between pedestrians and background in blurred and low-contrast circumstances. To validate the effectiveness of our model in nighttime pedestrian detection, we compare our \emph{EGCL} model with state-of-the-art pedestrian detectors on NightOwls dataset. These methods include ACF~\cite{dollar2014fast}, Checkerboards~\cite{zhang2015filtered}, Vanilla Faster R-CNN~\cite{ren2016faster}, Adapted Faster R-CNN~\cite{zhang2017citypersons}, RPN+BF~\cite{zhang2016faster}, SDS-RCNN\cite{brazil2017illuminating} and recent TFAN~\cite{wu2020temporal}.
The results in Table~\ref{tab:comparison_nightowls} show that \MR{our model outperforms other methods}, which reveals the advantage of our model in nighttime  detection over other methods. 

\begin{table}[!t]
\centering
\caption{Performance ( in terms of $MR^{-2}$) of our \emph{EGCL} and other methods on TJU-DHD-pedestrian dataset including two sub-datasets (lower is better).}
\renewcommand\arraystretch{1.5} 
\resizebox{1\linewidth}{!}{
\begin{tabular}{c| l |  c c c c}
\toprule
\multicolumn{1}{c|}{Subset}&\multicolumn{1}{c|}{Methods} &\multicolumn{1}{c}{\textbf{R}}& \multicolumn{1}{c}{\textbf{HO}}  & \multicolumn{1}{c}{\textbf{R+HO}} & \multicolumn{1}{c}{\textbf{ALL}} \\
\midrule
\multirow{5}*{TJU-Ped-campus}&RetinaNet~\cite{lin2017focal}&34.73&71.31&42.26&44.34\\
~&FCOS~\cite{tian2019fcos}&31.89&69.04&39.38&41.62\\
~&FPN~\cite{lin2017feature}&27.92&67.52&35.67&38.08\\
~&CrowdDet~\cite{chu2020detection}&25.73&66.38&33.65&35.90\\
~&CrowdDet+\emph{EGCL} (ours)&\textbf{24.84}&\textbf{65.27}&\textbf{32.39}&\textbf{34.87}\\
\midrule
\multirow{5}*{TJU-Ped-traffic}&RetinaNet~\cite{lin2017focal}&23.89&61.60&28.45&41.40\\
~&FCOS~\cite{tian2019fcos}&24.35&63.73&28.86&40.02\\
~&FPN~\cite{lin2017feature}&22.30&60.30&26.71&37.78\\
~&CrowdDet~\cite{chu2020detection}&20.82&61.22&25.28&36.94\\
~&CrowdDet+\emph{EGCL} (ours)&\textbf{19.73}&\textbf{60.05}&\textbf{24.19}&\textbf{35.76}\\
\bottomrule
\end{tabular}
}
\label{tab:sota-tju}
\end{table}

\subsubsection{TJU-DHD-pedestrian dataset}
In the last set of experiments, we compare our \emph{EGCL} with other methods on TJU-DHD-pedestrian dataset, which is mixed with daytime and nighttime images covering more challenging scenes for pedestrian detection. We compare our \emph{EGCL} with the state-of-the-art methods that have been evaluated previously on TJU-DHD-pedestrian dataset, including RetinaNet~\cite{lin2017focal}, FCOS~\cite{tian2019fcos} and FPN~\cite{lin2017feature}.  Specifically, we conduct experiments on the two subsets with different scenes separately, namely TJU-Ped-campus and TJU-Ped-traffic.
The results in Table~\ref{tab:sota-tju} show that our model achieves the best results on all subsets of both `campus' and `traffic' sub-datasets, which validates the effectiveness of our method. Note that only methods with reported results on this dataset are listed. Here CrowdDet is used as the baseline for our model to have a fair comparison. Our \emph{EGCL} \MR{performs better than CrowdDet on all metrics}.
\vspace{-5pt}

\section{Conclusion}
In this paper, we have presented the Exemplar-Guided Contrastive Learing (\emph{EGCL}) model for pedestrian detection. \emph{EGCL} learns a feature transformation module by contrastive learning to project the initial feature space into a new feature space, in which the semantic distance between pedestrians is minimized to eliminate the appearance diversities of pedestrians while the semantic distance between pedestrians and background is maximized. Extensive experiments on both daytime and nighttime pedestrian detection validate the effectiveness of the proposed \emph{EGCL}.

\section*{Acknowledgment}
This work was supported in part by the NSFC fund (U2013210, 62006060, 62176077), in part by the Guangdong Basic and Applied Basic Research Foundation under Grant (2019Bl515120055, 2021A1515012528, 2022A1515010306), in part by the Shenzhen Key Technical Project under Grant 2020N046, in part by the Shenzhen Fundamental Research Fund under Grant (JCYJ20210324132210025), in part by the Shenzhen Stable Support Plan Fund for Universities (GXWD20201230155427003-20200824125730001, GXWD20201230155427003-20200824164357001), and in part by the Medical Biometrics Perception and Analysis Engineering Laboratory, Shenzhen, China.


\ifCLASSOPTIONcaptionsoff
  \newpage
\fi



%
\bibliographystyle{IEEEtran}
\bibliography{main.bbl}
\end{document}